\documentclass[sn-mathphys,Numbered]{sn-jnl}

\usepackage{graphicx}%
\usepackage{multirow}%
\usepackage{amsmath,amssymb,amsfonts}%
\usepackage{amsthm}%
\usepackage{mathrsfs}%
\usepackage[title]{appendix}%
\usepackage{xcolor}%
\usepackage{textcomp}%
\usepackage{manyfoot}%
\usepackage{multirow}%
\usepackage{booktabs}%
\usepackage{algorithm}%
\usepackage{algorithmicx}%
\usepackage{algpseudocode}%
\usepackage{listings}%
\usepackage{xcolor}%

\usepackage[numbers]{natbib} 
\usepackage{bibunits} 

\defaultbibliography{sn-article}

\renewcommand{\abstractname}{Summary}

\begin{document}

\begin{bibunit}[unsrt]


\title[Article Title]{Explainable deep learning improves human mental models of self-driving cars}


\author*[1]{\fnm{Eoin M.} \sur{Kenny}}\email{ekenny@mit.edu}

\author[2]{\fnm{Akshay} \sur{Dharmavaram}}

\author[2]{\fnm{Sang Uk} \sur{Lee}}

\author[2]{\fnm{Tung} \sur{Phan-Minh}}

\author[2]{\fnm{Shreyas} \sur{Rajesh}}

\author[2]{\fnm{Yunqing} \sur{Hu}}

\author[2]{\fnm{Laura} \sur{Major}}

\author[2,3]{\fnm{Momchil S.} \sur{Tomov}}\email{momchil.tomov@motional.com}
\equalcont{These authors contributed equally to this work.}

\author[1,4]{\fnm{Julie A.} \sur{Shah}}\email{julie\_a\_shah@csail.mit.edu}
\equalcont{These authors contributed equally to this work.}

\affil[1]{\orgdiv{Computer Science \& Artificial Intelligence Laboratory (CSAIL)}, \orgname{Massachusetts Institute of Technology}, \orgaddress{
\city{Cambridge}, 
\state{MA}, \country{USA}}}

\affil[2]{ \orgname{Motional AD Inc.}, \orgaddress{\city{Boston}, 
\state{MA}, \country{USA}}}

\affil[3]{\orgdiv{Department of Psychology and Center for Brain Science}, \orgname{Harvard University}, \orgaddress{\city{Cambridge}, 
\state{MA}, \country{USA}}}

\affil[4]{\orgdiv{Department of Aeronautics and Astronautics}, \orgname{Massachusetts Institute of Technology}, \orgaddress{
\city{Cambridge}, 
\state{MA}, \country{USA}}}


\abstract{
Self-driving cars increasingly rely on deep neural networks to achieve human-like driving~\cite{tomov2025treeirl,phan2023driveirl,marcu2023lingoqa}. 
The opacity of such black-box planners makes it challenging 
to accurately anticipate when they will fail~\cite{nikolaidis2012human,major2020expect,paleja2021utility}, with potentially catastrophic consequences~\cite{Titcomb2019,Fang2023,templeton2024waymo}.
While research into interpreting these systems has surged, most of it is confined to simulations or toy setups due to the difficulty of real-world deployment~\cite{kenny2023towards,atakishiyev2024explainable}, leaving the practical utility of such techniques unknown.
Here, we introduce the Concept-Wrapper Network (CW-Net), a method for faithfully explaining the behavior of machine-learning-based planners that causally grounds their reasoning in human-interpretable concepts without sacrificing performance. 
We deploy CW-Net on a real self-driving car and show that the resulting explanations improve the human driver's mental model of the vehicle, allowing them to better predict its behavior, particularly in surprising situations.
This demonstrates that explainable deep learning integrated into self-driving cars can be both understandable and useful in a realistic deployment setting.
We anticipate our method could be applied to other safety-critical systems, such as autonomous drones and robotic surgeons, as well as to other architectures, such as end-to-end learning systems and vision-language-action models.
Overall, our study establishes a deployment-validated pathway to interpretability for autonomous agents, which could help make them more transparent and safe.
}

\keywords{Self-driving cars, Interpretable ML, Explainable AI, Mental models, Human–computer interaction, Situational awareness}

\maketitle

\section{Introduction}
\label{sec1}

There are hundreds of companies developing autonomous vehicle (AV) technology globally~\cite{badue2021self}, promising to revolutionize transportation for everyone.
The industry currently spans two segments: fully autonomous ride-hail systems and consumer vehicles with driver-assistance~\cite{SAEJ3016_2021}. 
In consumer vehicles, machine learning (ML) solutions have markedly improved the technology, yet they still require human intervention in unusual or challenging situations where learned planners may not determine the correct action~\cite{xing2021toward}. 
As driving is safety-critical, such infrequent failures matter, making it essential that the human driver is able to anticipate and be prepared for such situations~\cite{pereira2020challenges}. 
However, the opaque nature of ML planners makes it challenging to interpret and communicate the causes of their decisions, hampering the ability of human drivers to understand and predict AV behavior while achieving real-time situational awareness~\cite{kuznietsov2024explainable,arfini2023design,atakishiyev2024incorporating}.


Lack of effective communication between the AV and the human driver has contributed to multiple high-profile incidents, some resulting in fatalities~\cite{Titcomb2019,Fang2023,templeton2024waymo}, highlighting the urgent need to make ML planners interpretable~\cite{atakishiyev2024explainable}.  
Previous studies have sought to address this using surveys and simulated scenarios~\cite{koo2015did,koo2016understanding,wiegand2020d,wang2021human,schneider2021increasing,schneider2021explain,omeiza2021towards,zemni2023octet}, a human driver emulating the AV~\cite{kim2023and,schneider2023don}, or language models providing rationales for the driving policy in natural language~\cite{marcu2023lingoqa,wang2025alpamayo}. 
However, these studies were theoretical, did not provide causally faithful explanations, were only evaluated in simulation, or did not convincingly show the practical utility of the explanations to end users.
This leaves open the question of how to provide explanations that are understandable, useful, and faithful to the decision-making process of the AV in a realistic setting.

To answer this question, we scale up our work on interpretable-by-design deep reinforcement learning~\cite{kenny2023towards} using motifs from the literature on concept-bottleneck models~\cite{yuksekgonulpost} to propose the Concept-Wrapper Network (CW-Net). 
CW-Net grounds the reasoning of a black-box ML planner in human-interpretable concepts, such as \textit{``Approaching stopped vehicle''} or  \textit{``Close to cyclist''}.
This method is rooted in case-based reasoning, a classical artificial intelligence (AI) approach~\cite{leake1996case,keane2019case,sormo2005explanation,kenny2019twin} inspired by cognitive models of human reasoning and memory~\cite{schank1983dynamic}. 
CW-Net can be applied to arbitrary pretrained deep neural networks, does not require retraining from scratch, and does not degrade the performance of the original black-box ML planner.
Since the inferred concepts are the sole input to the final decision-making module of CW-Net and therefore directly determine AV behavior, we refer to them as causally faithful. 
Notably, our approach contrasts with popular post-hoc explanation methods~\cite{lundberg2017unified,ribeiro2016should}, which are applied to pretrained models and do not, by construction, guarantee faithfulness to the model decision process~\cite{rudin2019stop}.


We apply CW-Net to an ML planner trained to imitate human driving behavior using inverse reinforcement learning~\cite{tomov2025treeirl,phan2023driveirl}.
We replace the final (reward) layer of the pretrained deep neural network with a concept classifier, followed by a new reward layer.
We then jointly train the classifier and the new reward layer to predict scenario types and driving decisions, respectively, without modifying the rest of the network. 
Evaluation on a large-scale benchmark~\cite{karnchanachari2024towards} confirms that CW-Net is able to classify concepts without compromising driving behavior. 
To study the utility of the explanations, we deploy CW-Net on a real self-driving car with a safety driver in a semi-naturalistic study~\cite{heim2025lab2car}. 
We demonstrate three situations in which the driver's initially inaccurate mental model is subsequently improved by the CW-Net explanations, which in turn allows the driver to more accurately predict AV behavior.
We then observe similar mental model improvement in larger online studies simulating these scenarios.
Lastly, we link these results to more complex real-world situations by deploying CW-Net on public roads in Las Vegas. 
We use these new data in a large online study ($N = 100$) to demonstrate that improvements in mental model goodness lead to improved predictive ability and situational awareness. 
Overall, our work demonstrates how explainable deep learning can help users of an advanced autonomous system improve their mental model of the system and better anticipate its future behavior.

\section{Black-box planner}

\begin{figure}[h!]
\centering
  \includegraphics[width=\textwidth,trim={15 90 200 80},clip]{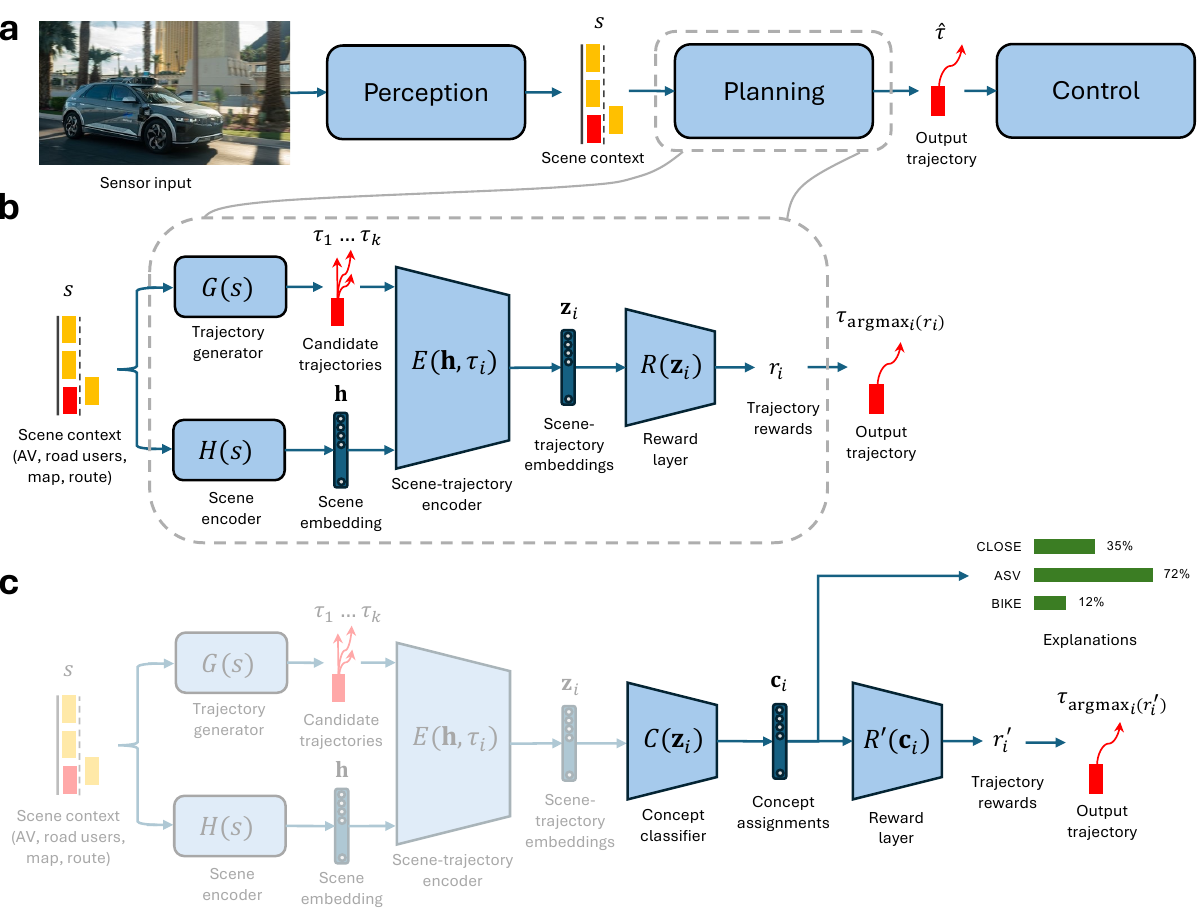}
  \caption{
  \textbf{Planner architecture}, adapted with permission from Tomov et al. (2025)~\cite{tomov2025treeirl}. 
  \textbf{a.} Autonomous vehicle stack.
  Sensory input is processed by the perception module to generate scene context $s$. 
  The planning module processes $s$ to compute trajectory $\hat{\tau}$, which is followed by the control module. 
  \textbf{b.} Black-box ML planner. 
  Here, $s$ is fed to trajectory generator $G$, which produces candidate trajectories $\{\tau_1 \dots \tau_k\}$, and scene encoder $H$, which produces a scene embedding $\mathbf{h}$.
  These are fed into encoder $E$, which produces scene-trajectory embeddings $\mathbf{z}_i$ for each $\tau_i$, which are in turn fed into the reward layer $R$. 
  $R$ computes a reward $r_i$ for each trajectory. 
  The model is trained to output higher rewards for trajectories closer to the the ground-truth human trajectories. 
  \textbf{c.} CW-Net. Identical to \textbf{b.}, except $\mathbf{z}_i$ is fed to a concept classifier $C$, which produces concept assignments $\mathbf{c}_i$. 
  These are fed to a new reward layer $R'$ to produce rewards $r'_i$.
  In parallel, $\mathbf{c}_i$ is processed to generate the explanations. 
  The reward layer is trained to prefer the same trajectories as the black-box ML planner, while the concept layer is supervised with ground-truth scenario labels. 
  The weights of the faded components are frozen during training.
  \texttt{CLOSE}, \textit{``Close to another vehicle''}. 
  \texttt{ASV}, \textit{``Approaching stopped vehicle''}. \texttt{BIKE}, \textit{``Close to cyclist''}.
  }
  \label{fig:architecture}
\end{figure}

We focus on the planning module of the AV stack (Figure~\ref{fig:architecture}a), which takes as input a scene context $s$ and outputs a trajectory $\hat{\tau}$. Here, $s$ is a symbolic object-oriented representation of the scene, computed by the perception module, while $\hat{\tau}$ is the trajectory that the subsequent controller module should follow. 
We use a deep neural network architecture~\cite{tomov2025treeirl,phan2023driveirl} consisting of a scene encoder $H(s) \rightarrow \mathbf{h}$ and a trajectory generator $G(s) \rightarrow \{\tau_1 \dots \tau_k\}$, followed by a scene-trajectory encoder $E(\mathbf{h}, \tau_i) \rightarrow \mathbf{z}_i$ and a final reward layer $R(\mathbf{z}_i) \rightarrow r_i$ 
(Figure~\ref{fig:architecture}b). 
$H$ computes a scene embedding $\mathbf{h}$ and $G$ computes a set of $k$ candidate trajectories $\{\tau_1 \dots \tau_k\}$. 
Those are combined in $E$ to compute an embedding $\mathbf{z}_i$ for each trajectory $\tau_i$. 
Finally, $R$ computes an estimated reward $r_i$ for each trajectory $\tau_i$, quantifying how human-like it is. 
In other words, $r_i$ is higher when $\tau_i$ is more similar to how a human would drive in this situation. 

During inference, the trajectory with highest reward is selected on each iteration:

\[
 \hat{\tau} = \underset{\tau \in { \{\tau_1 . . .  \tau_k} \} }{\arg\max} R(E(\mathbf{h},\tau))
\]

We use inverse reinforcement learning to train the planner on 80 hours of human expert driving~\cite{tomov2025treeirl,phan2023driveirl}. 

\section{Planning over human-friendly concepts}

To make the ML planner interpretable, we modify the planner architecture to additionally provide explanations for its behavior~\cite{kenny2023towards}.  
Our working definition of explainable AI follows Gunning \& Aha~\cite{gunning2019darpa}, who define it as \textit{`AI systems that can explain their rationale to a human user, characterize their strengths and weaknesses, and convey an understanding of how they will behave in the future.'} 

Specifically, we replace $R$ with a concept classifier $C(\mathbf{z}_i) \rightarrow \mathbf{c}_i$, followed by a new reward layer $R'(\mathbf{c}_i) \rightarrow r'_i$ (Figure~\ref{fig:architecture}c). 
The concept classifier $C$ computes a logit vector $\mathbf{c}_i$ that is passed through a softmax and/or sigmoid layer that, in turn, assigns probabilities to different human-interpretable concepts. 
Since $R'$ computes trajectory rewards from $\mathbf{c}_i$, the final decisions are based solely on these concept assignments and hence they constitute a causally faithful explanation. 
The rest of the network remains the same.

Similarly to the black-box planner, trajectories are selected according to:

\[
\hat{\tau} = \underset{\tau \in { \{\tau_1 . . .  \tau_k} \} }{\arg\max} R'(C(E(\mathbf{h},\tau)))
\]

We train CW-Net to jointly predict concept labels and mimic the driving decisions of the black-box planner. 
Specifically, $\mathbf{c}_i$ is supervised with 
labels 
corresponding to types of scenarios, such as \textit{``Approaching stopped vehicle''} or \textit{``Close to cyclist''} (see Section~\ref{methods:concept_details} for a full list of concepts). 
This ensures that CW-Net assigns a unique interpretable concept to each unit in $\mathbf{c}_i$. 
At the same time, $r'_i$ is supervised with trajectories selected by the black-box planner using a cross-entropy loss.
During training, the rest of the deep neural network ($H$, $G$, and $E$) is kept frozen (see Section~\ref{methods:training} for more details).
We focus our study on this particular ML planner architecture, noting that prior work indicates CW-Net would generalize effectively to other architectures~\cite{kenny2023towards}.

\section{Classifying concepts}

As a baseline, we first evaluated the black-box planner (without CW-Net) using closed-loop simulations with the nuPlan simulator on the nuPlan dataset~\cite{karnchanachari2024towards} (Table~\ref{tab:nuplan}). 
Overall, the results were competitive with the top submissions to the nuPlan challenge, although performance was slightly lacking when starting from a stop (see Section~\ref{methods:simulation}).
This suggests that there is room for improvement and, importantly, opportunities to study explanations of undesirable behavior.

We then evaluated the driving performance of CW-Net wrapped around the black-box planner (Table~\ref{tab:nuplan}). 
The results were equivalent, with less than 1\% difference across all metrics, confirming that our method did not degrade driving performance. We also evaluated concept classification on held-out datasets (Tables~\ref{table:concept500k},~\ref{table:concept3M}). 
Mean accuracy was 54\%, with 23\% precision, 77\% recall, and an F1 score of 0.31 (see Section~\ref{methods:concept_seperation}).  
Overall, these results indicate that CW-Net can be used to ground
the decision making of high-performance ML planners in human-interpretable
concepts, without sacrificing driving performance.


\section{Mental model improvement in deployment}
\label{sec:seminaturalistic}

Our central hypothesis is that the explanations from CW-Net would improve the human driver's mental model of the AV and, by extension, their situational awareness. 
This would be particularly salient in surprising situations, which is when explanations are most useful~\cite{foster2013surprise}. 
Specifically, the explanations should improve the driver's ability to understand and address the reasons for AV failures, and also to predict its future behavior. 

\begin{figure}[t!]
\centering
  \includegraphics[width=\textwidth]{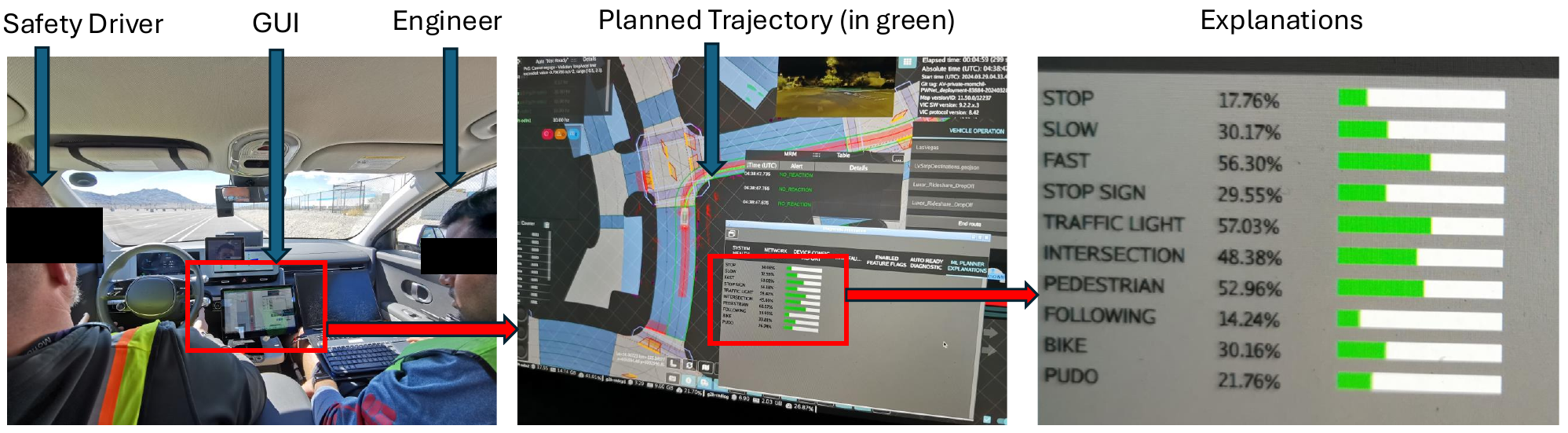}
  \caption{
  \textbf{Deployment setup.} 
  A safety driver, a support engineer, and a researcher were present.
  The safety driver drove the AV manually between road tests, engaged self-driving mode at the start of each test, monitored AV performance during the test, and took over in case of unsafe driving.
  The support engineer deployed CW-Net and set scenario destinations. 
  The researcher directed testing. 
  The dashboard included a map with overlaid object detections ($s$) from the perception module and the output trajectory ($\hat{\tau}$). 
  Explanations $\mathbf{c}_{\hat{i}}$ from CW-Net were shown as percentages for easier interpretation~\cite{gigerenzer1995improve}.
  }
  \label{fig:setup}
\end{figure}

To test this hypothesis, we deploy CW-Net on a real AV on a private track using the Lab2Car wrapper~\cite{heim2025lab2car} and observe how safety drivers react to surprising events and the corresponding CW-Net explanations (Figure~\ref{fig:setup}). 
These situations were not planned, but instead occurred naturally, with minimal intervention from the researchers, who only dictated high-level plans for each day.
This semi-naturalistic study allowed us to assess the utility of the explanations in naturally occurring surprising situations.
To measure AV predictability, we record the drivers' ability to make counterfactual predictions about AV behavior in these situations. 
To note the drivers' mental models, together with their predictive ability, we additionally record their think-aloud thoughts before and after considering CW-Net explanations~\cite{hoffman2023measures}. 
We focus on concepts that relate to other road users and can be easily tested counterfactually (see Section~\ref{methods:concept_details}).

\label{sec:deploy}

\subsection{Unexpected stopping for nearby vehicles}
\label{sec:close}

\begin{figure}[h!]
\centering
 \includegraphics[width=\textwidth]{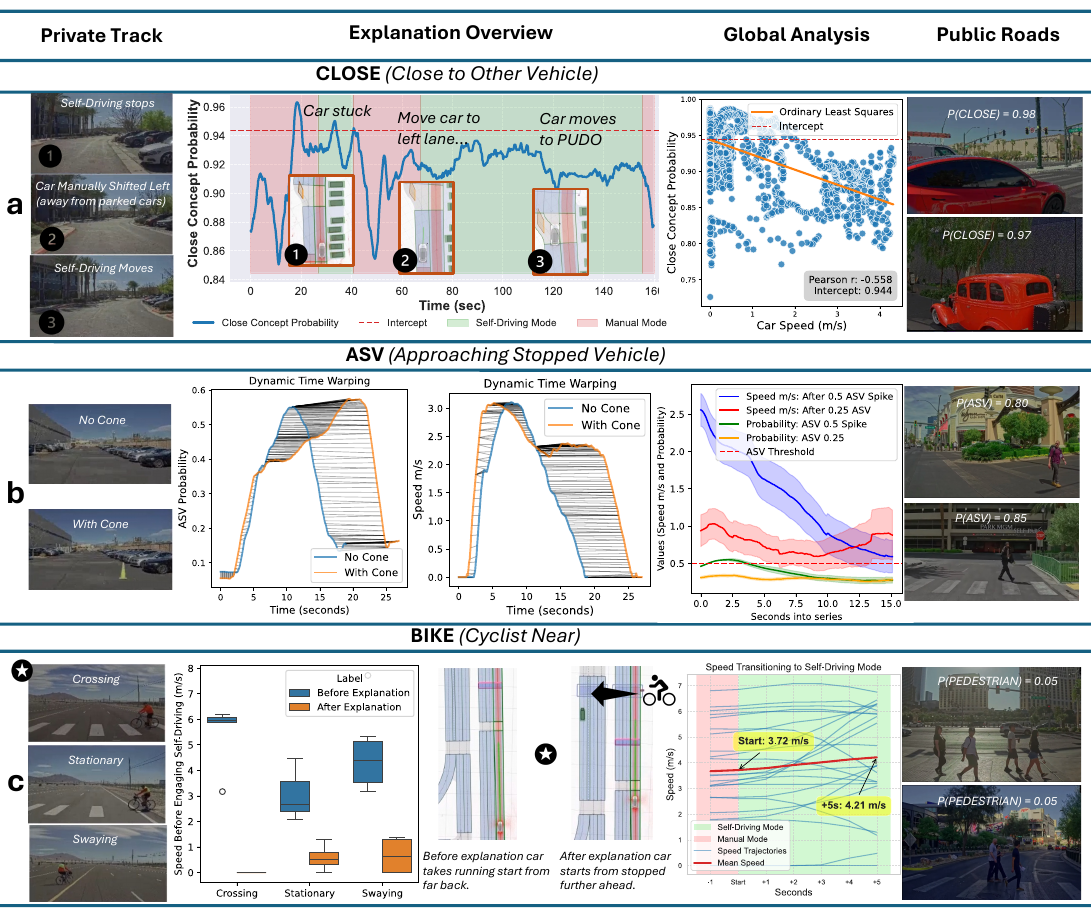}
   \caption{
    \textbf{Results}: 
    \textbf{a.} The \texttt{CLOSE} concept activated when the car got stuck next to parked vehicles.
    The driver initially thought the pickup/drop-off area was the cause, but the explanation suggested that it was the nearby vehicles. 
    When the driver engaged self-driving away from the park vehicles, activation of the \texttt{CLOSE} concept decreased and the AV started moving again, counter to driver’s initial mental model and consistent with the explanations.
    Across tests, \texttt{CLOSE} correlated with speed and the intercept accurately predicted this event.
    \textbf{b.} The \texttt{ASV} concept activated when the car stopped next to a traffic cone. 
    The driver initially thought the cone was the cause, but the explanation suggested that the AV was hallucinating a stopped vehicle.
    When we removed the cone in a counterfactual test, the same phantom braking and concept activation occurred.
    Across tests, \texttt{ASV} correlated with reductions in speed when spiking above 0.5 probability.
    \textbf{c.} The \texttt{BIKE} concept failed to activate in our first round of tests with a cyclist, but the car always stopped safely for the cyclist.
    After observing the explanation, the driver engaged self-driving from slower speeds in a second round of tests.
    Follow-up analyses revealed that the AV stopped for the cyclist due to backup safety mechanisms unrelated to CW-Net, which indicates that the driver's increased level of caution was appropriate.
    \textbf{a-c.} (Right column) Naturalistic scenarios from public-road deployment, analogous to the scenarios observed on the private track. 
    We used the \texttt{PEDESTRIAN} concept instead of the \texttt{BIKE} concept as it behaved similarly and pedestrians were more frequently encountered on public roads.
    Likewise, we used pedestrians in place of the traffic cone in our \texttt{ASV} tests as there were no cones encountered on the public roads.
    Standard error of the mean and 3-second rolling averages are shown in relevant plots.
    }
  \label{fig:results}
\end{figure}

We observed that the AV repeatedly came to a stop shortly before a pedestrian pickup/drop-off zone (Figure~\ref{fig:results}a). 
The driver's intuition was that the car stopped because of the pickup/drop-off zone, 
but the explanations indicated that the planner stopped because it detected that it was \textit{``Close to another vehicle''} (the \texttt{CLOSE} concept). 
To test this hypothesis, the driver manually moved the car farther from the parked cars. 
At this point, the probability of \texttt{CLOSE} decreased and the AV began moving again, 
thus supporting the alternative hypothesis. 
A full timeline of events is detailed in Figure~\ref{fig:results}a. 
We fitted the intercept of the \texttt{CLOSE} probability against the speed of the AV globally and found it accurately predicts stopping and starting for this event. 


\subsection{Hallucinating a stopped vehicle ahead}

At another location, the AV would reliably come to a stop next to a traffic cone (Figure~\ref{fig:results}b). 
The driver's initial mental model was that the cone was responsible for the phantom brake.
However, the \textit{``Approaching stopped vehicle''} (\texttt{ASV}) concept peaked shortly before the car stopped. 
This suggested an alternative hypothesis, that the planner matched the current situation with training scenarios labeled \texttt{ASV}, 
which in turn promotes stopping behavior associated with such scenarios.
As a counterfactual test, the cone was removed. 
The AV exhibited the same stopping behavior at the same location, along with similar \texttt{ASV} probability and speed profiles ($L_2$ similarities of 7.37 (1.12 original version) and 1.6 between the respective time-warped profiles, compared to an average $L_2 > 200$ for random events), thus supporting the alternative hypothesis.
Note that although there was no vehicle in front of the AV, the explanation is causally faithful to the underlying planner and explains why it stopped (namely, because it incorrectly detected a stopped vehicle). 
Figure~\ref{fig:results}b illustrates a global analysis of \texttt{ASV}, showing it to be a powerful predictor of braking.

\subsection{Reacting safely to cyclist}

Finally, we tested the ability of the AV to 
stop safely for cyclists (Figure~\ref{fig:results}c). 
For each test, the driver engaged self-driving mode while approaching a cyclist.
The driver was instructed to engage self-driving from a speed at which they felt safe, since this determines the subsequent speed of the AV. 
During the initial tests, the AV reliably stopped for the cyclist. 
However, the \texttt{BIKE} concept maintained a low probability throughout each test ($<$ 1\%), indicating that CW-Net was failing to detect the cyclist.
Over time, the driver became aware of the concept reading and gradually increased their 
caution by initiating self-driving from slower speeds. 
A post-hoc  analysis revealed that, although the perception system detected the cyclist, the ML planner was not configured to consume inputs for cyclists.
As a result, it chose unsafe trajectories which would have collided with the cyclist. 
In reality, the AV stopped due to a built-in safety backup system, which commands a brake if collision is imminent. 
This indicates that the increased caution dictated by the driver's updated mental model was warranted.


\section{Simulation studies}

The semi-naturalistic study above illustrates how CW-Net explanations can improve the human driver's mental model of the AV in surprising real-world situations, leading to better understanding and predictability of AV behavior.
To validate that these results replicate in larger populations and generalize to naturalistic scenarios, we conduct several larger follow-up studies (Figures~\ref{fig:studies_overview}-\ref{fig:concepts_user}). 
First, we show that in simulations of the real-world scenarios in Section~\ref{sec:seminaturalistic}, CW-Net explanations consistently improve mental models and predictions for both experts and non-experts alike. 
Second, we collect data from CW-Net in naturalistic real-world scenarios on public roads and show that the explanations improve situational awareness -- an indirect measure of mental model goodness~\cite{endsley1995measurement} -- in surprising situations, without degrading it in unsurprising situations.


\subsection{Mental models and predictability}
\label{sec:mm_elicitation_studies}

The purpose of the first study is to (1) demonstrate that the effects observed on the road reproduce in larger populations, and (2) establish a link between direct measures of mental model goodness and performance on downstream tasks --- such as predicting AV behavior --- so that the latter can be used as a proxy for the former. 
We conducted an online survey in which we showed participants replays of the scenarios from the private track (front camera recordings with overlaid CW-Net explanations; see Figure~\ref{fig:analogous_scenarios}).

For each scenario, we asked participants to choose the reason for the AV's behavior that best aligns with their current beliefs (i.e., \textit{``Why did the AV do that?''}) and to make a counterfactual prediction (i.e., \textit{``What would the AV do if ...?''}). 
Following Hoffman et al.~\cite{hoffman2023measures}, we refer to the former as the \textit{nearest-neighbor task} -- since it forces participants to pick the nearest explanation to their beliefs -- and the latter as the \textit{prediction task}.
Both tasks consist of a multiple-choice question followed by a confidence score. 
The prediction task additionally includes a free-form text response where participants describe the reasons for their prediction. 
Both directly probe the participants' mental models, while the prediction task itself also serves as a proxy performance metric for mental model goodness. 
Importantly, we collected responses before and after observing the explanations. 
This within-participant design mirrors the experience of the drivers during the real-world tests (Figures~\ref{fig:studies_overview}, \ref{fig:analogous_scenarios}).
The study was conducted with experts ($N = 9$ Motional drivers/engineers) and non-experts ($N=30$ pseudo-random participants from Prolific.com).

On the nearest-neighbor task, CW-Net explanations improved mental models for almost all participants (8/9 experts and 27/30 non-experts; Figure~\ref{fig:MM_combined}). 
Free-form text responses were initially similar to the initial beliefs of the safety drivers during the on-road tests ($p < 10^{-9}$) and then shifted towards their final beliefs ($p < 0.0002$) and the ground-truth reasons for AV behavior  ($p < 10^{-5}$; exact binomial tests for non-experts; Figure~\ref{fig:llm_confusion_matrix__and__correlation_MM_heatmap}A,B). 
The distribution of mental model updates on both tasks was similar across both groups ($\beta = 0.04, p = 0.8$, ordinary least-squares regression; Figure~\ref{fig:llm_confusion_matrix__and__correlation_MM_heatmap}C).
Importantly, mental model improvement on the nearest-neighbor task correlated with improvements in prediction accuracy ($\beta=2.02\pm0.87, p=0.02$ for experts, $\beta=9.86\pm2.07, p<0.001$ for non-experts; linear mixed-effects models [LMEs]; Figure~\ref{fig:MM_cor_pred}, left). 
A similar effect was observed for the free-form text responses ($\beta=1.70\pm0.91, p=0.06$ for experts, $\beta=5.03\pm1.23, p<0.001$ for non-experts; LMEs; Figure~\ref{fig:MM_cor_pred}, right).
Together, these results support the conclusion that CW-Net explanations can reliably improve mental models of the AV and that prediction performance can serve as a reliable proxy for mental model goodness.

\subsection{Explanations and situational awareness}

To verify that the effects observed on the private track generalize to more complex, naturalistic scenarios, we deployed CW-Net on public roads in Las Vegas (Figures~\ref{fig:analogous_scenarios}, \ref{fig:concepts_user}, \ref{fig:results}-right). 
Due to safety reasons and the experimental nature of the ML planner,
the safety driver operated the AV in `manual' mode for several hours, with CW-Net running in the background. 
This allowed us to collect naturalistic scenarios analogous to those encountered on the private track~\cite{FAA2011FlightTest,schneider2023don} (Figure~\ref{fig:analogous_scenarios}). 
We use replays of those scenarios to conduct a large-scale online study ($N = 100$) within the Situation Awareness Global Assessment Technique (SAGAT) framework tailored for explainable AI~\cite{endsley1995measurement,sanneman2022situation}.


The SAGAT framework is the gold standard for evaluating situational awareness in complex, dynamic tasks~\cite{endsley1995measurement,sanneman2022situation}. 
In contrast to prior benchmark applications of SAGAT in driving research, which have been conducted almost exclusively within controlled driving-simulator environments \cite{ma2005situation,scholtz2005implementation}, we evaluate situational awareness on replays drawn from real-world public-road operation of the AV, addressing long-standing concerns about the ecological validity of simulator-only assessments.
It measures the situational awareness  of a human operator by freezing a scenario, querying the operator about their \textit{perception} (i.e., \textit{``What are the inputs to the AV?''}), \textit{comprehension} (i.e., \textit{``Why is the AV doing that?''}), and \textit{projection} (i.e., \textit{``What would the AV do if ...?''}) of the situation (Figure~\ref{fig:SAGAT_example_and_results}) and comparing their responses with the ground truth. 
As counterfactual predictive performance in Section~\ref{sec:mm_elicitation_studies} aligns with mental model goodness, we use projection in SAGAT as a proxy for mental model elicitation and goodness~\cite{hoffman2023measures}.
For each type of scenario and corresponding concept, we sample two instances in which AV behavior was surprising (e.g., stopping unnecessarily) and two corresponding instances in which it was unsurprising (Table~\ref{tab:non_surprising}), to verify CW-Net did not negatively impact situational awareness.
We use a between-participant design in which the experimental group received explanations from CW-Net, while the control group received placeholder explanations of AV kinematics to balance cognitive load~\cite{kenny2021explaining} (Figure~\ref{fig:concepts_user}).

We found that the CW-Net explanations significantly improved measures of
situational awareness for surprising events (Figure~\ref{fig:SAGAT_example_and_results}, left; Table~\ref{tab:survey_analysis}, top), with large effect sizes for perception (Cohen's d = 1.290, 95\% CI [0.857, 1.723]), comprehension (Cohen's d = 0.996, 95\% CI [0.578, 1.413]), and a medium effect size for projection (Cohen's d = 0.606, 95\% CI [0.203, 1.009]).
Conversely, explanations did not impact situational awareness for unsurprising events after Bonferroni correction (Figure~\ref{fig:SAGAT_example_and_results}; Table~\ref{tab:survey_analysis}). 
Together, these results show that the explanations from CW-Net are robust across various conditions and can reliably improve mental models in surprising situations, without significant negative effects in unsurprising situations. 
Moreover, they demonstrate pragmatic usage of the improved mental models.

\section{Conclusion}
\label{sec:discussion}
Our work shows how explainable deep learning can provide useful explanations for AVs in a real-world setting.
CW-Net achieves this by grounding the reasoning of a pretrained black-box ML planner in human-interpretable concepts that are directly used to make driving decisions. 
By revealing otherwise inaccessible information about the decision-making process of the AV in real time, CW-Net helps improve the human driver's mental model of the AV.
This, in turn, improves the driver's situational awareness and reveals limitations of the robotic system, helping the driver better anticipate its mistakes.
While we showcase CW-Net using a particular kind of classification-based ML planner architecture, the core idea can be similarly applied to other architectures, including end-to-end learning systems and vision-language-action models. Additionally, while our experimental setup assumes a human driver observing the explanations in real time --- an application more suited to advanced driver-assistance systems --- CW-Net could equally be applied to debugging and improving fully autonomous AV's.
Critically, whereas prior work on explanations and situational awareness for AV decision making has largely been confined to simulated or controlled scenarios~\cite{koo2015did,koo2016understanding,wiegand2020d,wang2021human,schneider2021increasing,schneider2021explain,omeiza2021towards,zemni2023octet,sanneman2022situation}, our deployment of CW-Net on public roads extends these findings to the environmental complexity of real-world driving, providing evidence of ecological validity and practical relevance for the AV industry.





Many systems involving human-robot interaction require real-time explanations, including AI wingmen, drone navigation systems, and robotic surgeons.
Similarly to AVs, many of these applications increasingly rely on deep learning, with a long tail of potentially catastrophic failure cases.
Indeed, many regulatory bodies have already made explainable AI a core component of their legislation, with AVs likely to follow suit as they are widely deployed with various users~\cite{atakishiyev2024explainable}.
As such, the success of CW-Net suggests that similar algorithms may prove essential for meeting the regulatory standards for deploying AVs, while building appropriate trust in the technology. 
In future work, it would be prudent to extend CW-Net to a larger set of concepts --- perhaps in an unsupervised manner to overcome the challenges of labeling --- and better cover the vast array of concepts relevant to AV settings.

\putbib
\end{bibunit}

\newpage

\begin{bibunit}[unsrt]

\section{Figure legends}
For initial submissions, we encourage authors to present the manuscript text and figures together in a single Word doc or PDF file, and for each figure legend to be presented together with its figure. However, when preparing the final paper to be accepted, we require figure legends to be listed one after the other, as part of the text document, separate from the figure files, and after the main reference list.

Each figure legend should begin with a brief title for the whole figure and continue with a short description of each panel and the symbols used. If the paper contains a Methods section, legends should not contain any details of methods. Legends should be fewer than 300 words each.

All error bars and statistics must be defined in the figure legend, as discussed above.

\section{Methods}
\label{methods}


\subsection{Architecture}
\label{methods:architecture}

The black-box ML planner uses a modified version of the DriveIRL architecture (Figure~\ref{fig:architecture}b)~\cite{tomov2025treeirl,phan2023driveirl}. 
For the trajectory generator $G$, we use a heuristic generator that produces 143 jerk-optimal trajectories to anchor waypoints  along the route. 
For the scene encoder $H$, we use the hierarchical vector transformer (HiVT)~\cite{zhou2022hivt} pretrained for multi-agent motion prediction. 
In addition to the scene embedding $\mathbf{h}$, this produces an additional 3 trajectories for the AV, for a total of $k = 146$ candidate trajectories. 
In the scene-trajectory encoder $E$, trajectories are encoded using a recurrent neural network (RNN) and then fed jointly with the scene embedding into a transformer layer which produces the scene-trajectory embeddings $\mathbf{z}_i$. 
The reward model $R$ is a multilayer perceptron (MLP).
In CW-Net (Figure~\ref{fig:architecture}c), the classifier $C$ and the new reward model $R'$ are MLPs.

We avoided testing other methods besides CW-Net because, at the time of writing, we are unaware of other works which are capable of modeling interpretable-by-design IRL systems.
Moreover, we broadly seek a general comparison of concept-based explanations (i.e., CW-Net) against no explanation (i.e., our control) to help generalizability of the results across the myriad of concept-based explainability techniques in the literature.

\subsection{CW-Net training}
\label{methods:training}

We used two datasets:
\begin{itemize}
    \item Dataset 1: 500,000 scenarios, 8 concept labels (Table~\ref{table:concept500k}), and
    \item Dataset 2: 3,000,000 scenarios, 10 concept labels (Table~\ref{table:concept3M}).
\end{itemize}

For a full list of the concept labels and their meanings, see Section~\ref{methods:concept_details}.
Each scenario was associated with 146 trajectories, thus giving between 70.5-423 million training data points for the concept classifier, each with multiple concept labels.
Our algorithm assumes CW-Net training has access to the original dataset used to train the black-box ML planner, along with annotated human-understandable concept labels for each of these data points. 
The annotations can be multi-label, meaning that one datum can be associated with as many concepts as desired or useful. 
The experiments with \texttt{CLOSE} and \texttt{ASV} concepts used models trained on Dataset 1. 
The experiments with \texttt{BIKE} and \texttt{PEDESTRIAN} concepts used models trained on Dataset 2.

During training, the parameters of the trajectory generator $G$, the scene encoder $H$, and the scene-trajectory encoder $E$ are frozen, and only the concept classifier $C$ and the new reward model $R'$ are trainable. 
Two separate losses are optimized jointly. 

First, a concept classification loss $\mathcal{L}_{\text{concept}}$ is used to train $C$ to predict the correct concept label(s). 
In our setting, this loss combines cross-entropy with binary cross-entropy for different concepts, depending on the semantics of the corresponding scenario types. 
For example, in Dataset 1, we use cross-entropy to model the steering concepts of the car (\texttt{LEFT}, \texttt{RIGHT}, and \texttt{STRAIGHT}), and the speed concepts (\texttt{STOPPED}, \texttt{SLOW}), while also using binary cross-entropy to predict the presence of other concepts such as \texttt{ASV}, \texttt{INTERSECTION}, and \texttt{CLOSE}. 
These losses are then averaged into one:
\[
\mathcal{L}_{\text{concept}} = \frac{1}{2k} \sum_{i=1}^{k} \left( \frac{1}{M_{\text{CCE}}} \sum_{j=1}^{M_{\text{CCE}}} \mathcal{L}_{\text{CCE}}(c_{i,j}, \hat{c}_{i,j}) + \frac{1}{M_{\text{BCE}}} \sum_{l=1}^{M_{\text{BCE}}} \mathcal{L}_{\text{BCE}}(c_{i,l}, \hat{c}_{i,l}) \right)
\]
where $M_{\text{CCE}}$ is the number of concepts modeled using categorical cross-entropy (e.g., steering and speed), and $M_{\text{BCE}}$ is the number of concepts modeled using binary cross-entropy (e.g., presence of features like \texttt{ASV}, \texttt{INTERSECTION}, and \texttt{CLOSE}). 
$c_{i,j}$ and $\hat{c}_{i,j}$ represent the true and predicted labels for the $j$-th concept under CCE for the $i$-th data point, while $c_{i,l}$ and $\hat{c}_{i,l}$ represent the true and predicted labels for the $l$-th concept under BCE for the $i$-th data point.
On Dataset 2, we take a different approach and model everything, including the speed concepts (\texttt{STOPPED}, \texttt{SLOW}, and \texttt{FAST}), with binary cross-entropy. 
In general, these parameters can be tuned to fit the task at hand.

Secondly, a cross-entropy loss $\mathcal{L}_{\text{trajectory}}$ is used to train the network to predict the correct trajectory, which we define as the original trajectory chosen by the black-box planner. 
Both losses are averaged:
\[
\mathcal{L}_{\text{total}} = \frac{1}{2} \left( \mathcal{L}_{\text{concept}} + \mathcal{L}_{\text{trajectory}} \right)
\]
A focal loss~\cite{lin2017focal} is applied to counteract data imbalances, just as in the original DriveIRL planner~\cite{tomov2025treeirl,phan2023driveirl}.
Computationally, our networks were trained on a large distributed setup using PyTorch Lightning.

\subsection{Concept separation}
\label{methods:concept_seperation}
When adding interpretability modules  post hoc,  as we have, there is the possibility that the network will not have learned to separate the concepts of interest, and thus fail to be able to predict them accurately~\cite{schrodi2024concept}.
In fact, we observed this in the experimental prototype we tested (see Table~\ref{table:concept3M}), when certain concepts such as \texttt{CLOSE} and \texttt{PEDESTRIAN} had poor precision and high recall, relatively speaking.
There are two important points to note here.
First, the better trained and more sophisticated an architecture is, the more it naturally learns to separate an impressive number of concepts in an unsupervised manner~\cite{templeton2024scaling,manning2020emergent} (often in the millions), so this is unlikely to be an issue for most companies with the flagship models in the future.
Secondly, even if CW-Net has not learned to separate certain concepts (e.g., red traffic lights vs. green ones), this could highlight the reason why the AV fails to act appropriately (e.g., stop vs. go) a given situation (e.g., because there are insufficient traffic light scenarios in the training data). Hence, from an explainability point of view, even concepts with low accuracy can be useful, as we demonstrate in the paper.

\subsection{Alternative architecture}
\label{methods:alternate_architecture}
Alongside our primary causal architecture illustrated in Figure~\ref{fig:architecture}, we also developed an alternative which gave post-hoc justifications for AV behavior (Figure~\ref{fig:parallel_arch}).
Specifically, we froze the weights of the pretrained black-box ML planner and trained a concept classifier head $C$ to work in parallel to to the reward layer $R$. Similarly to the causal architecture (Fig~\ref{fig:architecture}c), $C$ used the scene-trajectory embeddings $\mathbf{z}_i$ to classify the concepts.
This approach is beneficial because of its relative simplicity and accessibility, although the drawback is that it may be less faithful to the model's reasoning process, as the concept classifications are not directly used by the model to rank state-trajectory pairs.
However, there is ample evidence that such explanations are often faithful and capable~\cite{ribeiro2016should,scott2017unified}, so we include both as an option and demonstrate the utility of both. Specifically, this parallel architecture was used in the real-world scenarios with the \texttt{CLOSE} concept (Section~\ref{sec:close}).

\subsection{Concept details}
\label{methods:concept_details}

\textbf{Dataset 1} concepts were as follows:
\begin{itemize}
  \item \texttt{LEFT}, \texttt{RIGHT}, \texttt{STRAIGHT}: Classification of driving direction concepts, trained with cross-entropy loss. 
  For example, the concept \texttt{LEFT} represents training scenarios where the car was turning left.
  \item \texttt{STOPPED}, \texttt{SLOW}: Classification of car speed concepts, trained with cross-entropy loss. 
  The concept of e.g. \texttt{STOPPED} represents training scenarios where the car was stopped.
  \item \texttt{ASV} (approaching stopped vehicle): Scenarios in which the car was approaching stopped a vehicle. Trained with binary cross-entropy. 
  \item \texttt{INTERSECTION}: Scenarios in which the car was at an intersection. Trained with binary cross-entropy. 
  \item \texttt{CLOSE}: Scenarios in which the car was within 3 meters of another vehicle. Trained with binary cross-entropy. 
\end{itemize}

\noindent \textbf{Dataset 2} had the following concepts:
\begin{itemize}
  \item \texttt{SLOW}: Scenarios in which the car was driving at 1-2 m/s.
  \item \texttt{STOPPED}: Scenarios in which the car was stationary.
  \item \texttt{FAST}: Scenarios in which the car was driving faster than 2 m/s.
  \item \texttt{STOP SIGN}: Scenarios in which the car was close to a stop sign.
  \item \texttt{TRAFFIC LIGHT}: Scenarios in which the car was close to a traffic light.
  \item \texttt{INTERSECTION}: Scenarios in which the car was at an intersection.
  \item \texttt{Pedestrian}: Scenarios in which the car was close to a pedestrian.
  \item \texttt{FOLLOWING}: Scenarios in which the car was following another vehicle.
  \item \texttt{BIKE}: Scenarios in which the car was close to a cyclist.
  \item \texttt{PUDO} (pedestrian pickup/drop-off): Scenarios in which the car was in a pedestrian pickup/drop-off zone.
\end{itemize}


All concepts in Dataset 2 were trained with a binary cross-entropy loss. 

Although the AV was trained on 8-12 concepts (depending on the dataset), we focus our study on a subset of concepts (\texttt{CLOSE}, \texttt{ASV}, \texttt{BIKE}, \texttt{PEDESTRIAN}) because they are the only ones which relate to other road users and are easy to test counterfactually. The concept \texttt{FOLLOWS} involves other road users too, but no notable occurrences happened during real-world deployment. 

\subsection*{Simulation results}
\label{methods:simulation}

We tested our CW-Net model across the entire nuPlan validation dataset to see how its driving performance compares to the original black-box ML planner it was trained from.
The dataset is a large-scale planning benchmark for autonomous driving~\cite{karnchanachari2024towards} and measures how close a trained AV is to a human expert in $L_2$ distance, progress along the route, and safety (no collisions). 
In the black-box model, when following the lane or decelerating from high speed, the planner was able to make progress along the route ($> 93\%$ of human driving distance), while avoiding collisions ($> 90\%$ collision-free) and staying close to the ground-truth human expert trajectory ($< 1$ m displacement at 5 s).
Performance was worse when starting from a stop, with less progress ($74\%$ of human driving distance), more collisions ($81\%$ collision-free), and greater deviation from the human expert (1.2 m displacement at 5 s). 
Overall, the results showed our variation of the AV architecture had less than 0.01 $L_2$ difference to the original black-box agent on average across all measurements, and not meaningfully different, showing that it is possible to train our more interpretable model in Figure~\ref{fig:architecture} without sacrificing performance.
The full results are in Table~\ref{tab:nuplan}.

For concept accuracy verification, we used 5\% holdout data from our training datasets, the results are given in Table~\ref{table:concept500k} and Table~\ref{table:concept3M}.
Across both datasets, the mean accuracy was 0.54, precision 0.23, recall 0.77, and F1 Score 0.31.
Overall, the results suggested that CW-Net did not separate all concepts equally well, which suits our purposes as the explanations will highlight when and how this happens, and how it relates to driving performance, thus helping with mental model refinement (see Section~\ref{methods:concept_seperation}).
Notable results include an F1 score of 0.82 for detecting the \texttt{SLOW} concept, and $<$ 0.00 for detecting the \texttt{BIKE} concept, showing the latter is perhaps not well encoded or understood by the car.

\subsection*{Mental model elicitation studies}
These studies were designed to replicate the driver's experiences in the private track tests (Section~\ref{sec:seminaturalistic}) with a larger cohort, 
in order to help demonstrate the robustness of our findings regarding the drivers' mental models (Figure~\ref{fig:studies_overview}).
The same study was run separately with experts (other drivers and test engineers from Motional) and non-experts (randomly sampled users from Prolific.com).
Specifically, we were interested in testing the following hypotheses:

\begin{itemize}
    \item Hypothesis 1: The participants' responses \textit{before} observing the explanation (in the video replay) would be more consistent with the belief of the safety driver \textit{before} observing the explanation (in the car).
    \item Hypothesis 2: The participants' responses \textit{after} observing the explanation would be more consistent with the belief of the safety driver \textit{after} observing the explanation.
    \item Hypothesis 3: The participants' responses \textit{after} observing the explanation would be more consistent with the ground-truth reason for AV behavior.
    \item Hypothesis 4: Counterfactual prediction ability would correlate with user mental model goodness, the latter as measured by the nearest-neighbor tasks and the free-form text response of Hoffman et al.~\cite{hoffman2023measures}.
    \item Hypothesis 5: The two measures of mental model goodness would correlate with each other and have similar distributions across groups.
\end{itemize}






\paragraph{Design and materials}
We focused on the same three surprising events observed during the private track tests. The experiment was designed to measure mental models through a combination of nearest-neighor and prediction tasks, accompanied by confidence scores and free-form text responses, in which we could probe the participants' mental models~\cite{hoffman2023measures} (Figure~\ref{fig:studies_overview}).
First, users saw the respective video, and were asked to rate two possible reasons for AV behaviour, along with confidence scores (i.e., the nearest-neighbor mental model elicitation task).
Then, users were asked to make the same counterfactual prediction as the driver (i.e., the prediction task), along with a confidence score, and a free-form text rationale explaining their reasoning.
Then, users saw the same video with the explanation, and repeated the questions. This within-participant design mirrors the experience of the safety drivers during the on-road tests. 

\paragraph{Participants}
For the expert group, we recruited 9 safety drivers, test engineers, and test specialists from Motional (aged between 18-80, 8 male and 1 female). 
All participants volunteered to participate and were not paid.
For non-experts we sampled users sourced from \url{Prolific.com}, which is know for its high-quality user base.
Thirty users were randomly sampled U.S. citizens aged between 18-80, native English speakers, and an even number of male/female.
To ensure high-quality text responses, we paid users above the average rate with 15 USD per hour and stressed that they should only take the study if they were certain they understood the instructions.
The study received MIT IRB approval from \textit{Committee on the Use of Humans as Experimental Subjects} (COUHES) - \textbf{Exempt Id : E-5903}, start data July 1st 2024, end date August 31st 2026.

\paragraph{Metrics}

We used the nearest-neighbor task and the free-form text rationale as direct measures of mental models. 
For the nearest-neighbor task, we measured mental model goodness as a combination of choice accuracy and confidence. 
Specifically, we formalized mental model improvement as shifting from an incorrect to a correct belief (w.r.t. the ground-truth reasons for AV behavior), or increasing confidence in the correct belief, or decreasing confidence in the incorrect belief (Figure~\ref{fig:MM_combined}). 
For the free-form text rationale, we employed an LLM-as-a-judge (GPT-5) to determine which text response (before or after observing the CW-Net explanation) is closer to the ground-truth reason for AV behavior (see Supplemental Methods for details).
The LLM prompts were tuned over 3 iterations on the expert responses (Figure~\ref{fig:llm_confusion_matrix__and__correlation_MM_heatmap}A) and then evaluated once on the non-expert responses (Figure~\ref{fig:llm_confusion_matrix__and__correlation_MM_heatmap}B).
Mental model improvement was formalized as instances where the free-form response after observing the CW-Net explanation is closer to the ground truth, compared to the free-form response before observing the explanation.

We used the prediction task as a measure of downstream performance that relies on mental models, thus measuring them indirectly~\cite{endsley1995measurement,endsley2017direct}. 
We measured prediction improvement as a combination of choice accuracy and confidence in the same way as for the nearest-neighbor task (Figure~\ref{fig:MM_cor_pred}). 

To analyze the relationship between direct (nearest neighbor, free-form text rational) and indirect (prediction) measures of mental models, we employed a Linear Mixed-Effects Model (LME) to evaluate the effect of the mental model change (improve/worsened) on the prediction change variable (i.e., the delta in confidence change in the prediction), while controlling for individual participant variation as a random effect (Figure~\ref{fig:MM_cor_pred}). 
To collapse accuracy/confidence on a single scale for computing the confidence deltas, we simply flipped the sign of confidence values for inaccurate beliefs to negative.
All code for this analysis will be available upon publication (see code availability below).

\paragraph{Results}
We found evidence favoring all of our hypotheses. 

\begin{itemize}
    \item Hypothesis 1: The participants' \textit{initial} belief (before observing the explanation) was often times more similar to the safety driver's \textit{initial} belief ($p < 10^{-9}$, exact binomial test).
    \item Hypothesis 2: The participants' \textit{final} belief (after observing the explanation) was often times more similar to the safety driver's \textit{final} belief ($p < 0.0002$, exact binomial test).
    \item Hypothesis 3: The participants' \textit{final} belief (after observing the explanation) was often times more similar to the ground truth ($p < 10^{-5}$, exact binomial test).
    \item Hypothesis 4: There was a clear relationship between mental model category and predictive ability of users, with only nearest neighbor categories failing to reach significance with experts (see Figure~\ref{fig:MM_cor_pred}).
    \item Hypothesis 5: Interaction analysis revealed no significant difference in the relationship between nearest-neighbor score improvement and text rationale improvement across the two groups ($p = 0.842$; Figure~\ref{fig:llm_confusion_matrix__and__correlation_MM_heatmap}C), indicating a consistent underlying mechanism for both experts and non-experts.
    Non-experts demonstrated a significant positive correlation ($\beta = 0.27, p < 0.001$), suggesting that improved scores were strong predictors of improved mental model rationale in this group. 
    While experts exhibited a nearly identical positive coefficient ($\beta = 0.23$), the relationship did not reach statistical significance, likely due to a smaller sample ($p = 0.243$).
\end{itemize}

\subsection*{Public roads evaluation using SAGAT}
\label{methods:user_study}
We deployed CW-Net in manual mode on Las Vegas public roads to collect complex, naturalistic scenarios analogous to those already discovered during the private track tests (Figure~\ref{fig:analogous_scenarios}). 
We used these as materials for an online Situation Awareness Global Assessment Technique (SAGAT) study ($N=100$). 
This setup allows us to validate the robustness of the CW-Net explanations using a well-established rigorous framework (Figure~\ref{fig:studies_overview}).
We used a between-participant design in which we compared CW-Net explanations (experimental group) against baseline explanations describing speed and steering (control group)~\cite{kenny2021explaining} (Figure~\ref{fig:SAGAT_example_and_results}). 

\paragraph{Gathering materials}
We collected data for \texttt{ASV} (01:02:55), \texttt{CLOSE} (00:50:42), and \texttt{BIKE} ($>$3 hours). 
We labeled sequences where human driving mimicked the ML planner in surprising scenarios that resembled those already discovered in the private track tests (e.g., \texttt{CLOSE}: stuck beside vehicles; \texttt{ASV}: braking for hallucinations; Figure~\ref{fig:analogous_scenarios}). 
\texttt{PEDESTRIAN} replaced \texttt{BIKE} due to there being no naturally occurring cyclists in the new video data.
We sampled two surprising events per concept (6 total) and six corresponding ``unsurprising'' events to ensure explanations did not degrade situational awareness in regular driving (Table~\ref{tab:non_surprising}). 
Activation thresholds were 0.5 for \texttt{ASV}/\texttt{PEDESTRIAN} and 0.94 for \texttt{CLOSE}, derived from prior private track data (Figure~\ref{fig:results}).

Note the distribution of concept activations did not change significantly between the private-track tests and the public-road tests (Table~\ref{tab:wasser}), despite the tests being performed more than a year apart, on different AV's, with different software stacks, and under completely different conditions. This demonstrates the robustness and reliability of the algorithm.

\paragraph{Study design}
We used a between-participants design (N=100) assessing SAGAT perception, comprehension, and projection. Participants were split into experimental (CW-Net explanations) and control (speed/steering placeholder) groups. 
Attention checks based on material content and viewing times reduced the pool to 99 participants, 51 in the experimental, and 48 in the control.

\paragraph{Materials}
Stimuli included 13 videos (6 surprising, 6 unsurprising, 1 attention check). 
The experimental group saw concept activations, while the control saw speed and steering data (Figure~\ref{fig:concepts_user}). 
Following each video, a ``blackout'' screen presented six binary questions: four related to \textit{perception}, one to \textit{comprehension}, and one to \textit{projection}.

\paragraph{Participants}
We recruited gender-balanced U.S. residents (18+, native English speakers) via Prolific.com.
participants were paid \$12/hr. 
The study received MIT IRB approval from \textit{Committee on the Use of Humans as Experimental Subjects} (COUHES) - \textbf{Exempt Id : E-5903}, start data July 1st 2024, end date August 31st 2026.

\paragraph{Metrics}
We analyzed the average of each participant on each question type using two-tailed t-tests, splitting data by surprising vs. unsurprising events for each situational awareness dimension.

\paragraph{Results}
After collection and attention check filtering, we collected 99 responses out of the target 100. 
Results are displayed in Figure~\ref{fig:SAGAT_example_and_results}. 
In surprising events, explanations significantly improved situational awareness after Bonferroni correction, showing large effect sizes for perception (Cohen's $d = 1.290$) and comprehension ($d = 0.996$), alongside a medium effect for projection ($d = 0.606$). 
In unsurprising events, no significant differences occurred (perception $d = 0.085$, projection $d = -0.142$, comprehension $d = -0.514$), confirming that explanations provided benefit in anomalous situations without adversely affecting situational awareness during routine operations.
Additionally, we confirmed feature robustness by comparing concept distributions to prior private track tests using Wasserstein distance (Table~\ref{tab:wasser}), finding no meaningful changes.

\subsection{Data availability}
The data used for plotting this manuscript's figures are available at \url{https://github.com/EoinKenny/CW-Net-Autonomous-Driving}, with the exception of Tables~\ref{tab:nuplan}, \ref{table:concept500k}, and \ref{table:concept3M}.
The AV model weights are not available due to intellectual property restrictions.
The videos of CW-Net and corresponding explanations shown to participants in our studies are available on the project website at \url{https://tomov.github.io/CW-Net/}.

\subsection{Code availability}
The code for reproducing the manuscript's plots are available at \url{https://github.com/EoinKenny/CW-Net-Autonomous-Driving}.
Due to Motional intellectual property issues, the code for training the real-world AV used in the paper cannot be made available.
However, we have provided a code ocean capsule which reproduces the algorithm in another toy self-driving domain.



\paragraph{Acknowledgements} We thank Tom Bewley of JPMC AI research, Mycal Tucker of Anthropic, and Winthrop Gillis of Harvard for reviewing an early version of the manuscript.

\paragraph{Author contributions} E.M.K contributed to conceptualization of the research, model training, experimental evaluation, writing.
A.D., S.U.L., T.P.M., S.R., Y.H., and M.S.T. all contributed to the technical implementation of the algorithm in the self-driving car.
L.M. contributed to project organization, and writing.
M.S.T. and J.A.S. contributed to conceptualization of the research, project organization, and writing.

\paragraph{Competing interests} The authors declare no competing interests.

\paragraph{Additional information}
Supplementary information available at \url{https://github.com/EoinKenny/CW-Net-Autonomous-Driving}.
Correspondence and requests for materials should be addressed to Eoin M. Kenny.



\putbib
\end{bibunit}

\newpage

\section*{Extended Data}

\setcounter{figure}{0}
\setcounter{table}{0}

\renewcommand{\thefigure}{ED\arabic{figure}}
\renewcommand{\thetable}{ED\arabic{table}}

\begin{figure}[h!]
\centering
  \includegraphics[width=\textwidth,trim={130 280 160 100},clip]{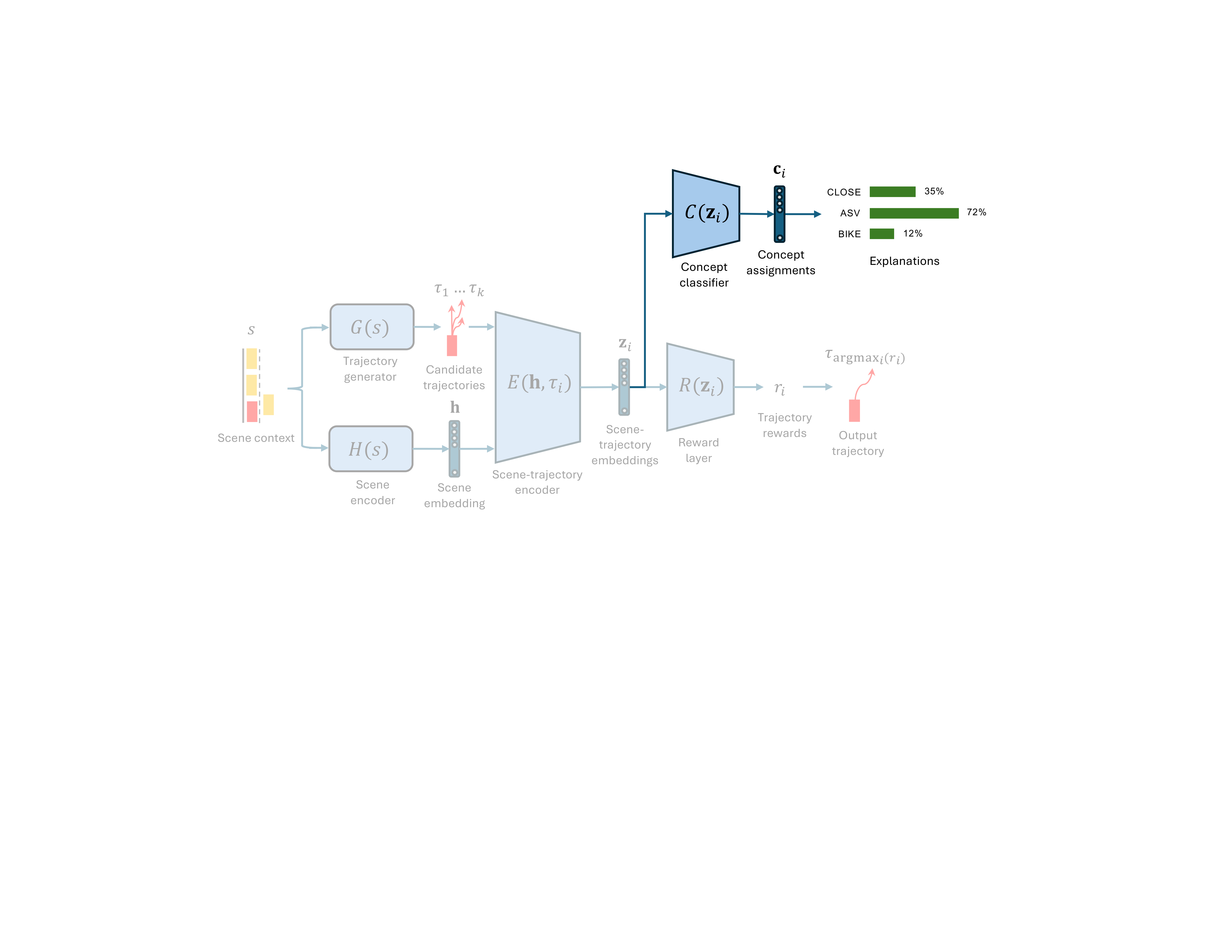}
  \caption{
  \textbf{Parallel architecture:} 
   This is identical to the black-box planner (Figure~\ref{fig:architecture}b; greyed out here), except the scene-trajectory embeddings are fed to the concept classifier $C$ in parallel to the (original) reward model $R$. The black-box part of the architecture (greyed out) is kept frozen.
   These concept classifications are then converted to probabilities (x100 to convert to percentages) and presented to the user.
  }
  \label{fig:parallel_arch}
\end{figure}



\begin{table}[!h]
\begin{tabular}{@{}llrr@{}}
\toprule
\multicolumn{1}{c}{Scenario type}                                                                                                          & \multicolumn{1}{c}{Metric}                                      & \multicolumn{1}{c}{CW-Net} & \multicolumn{1}{c}{ML planner} \\ \midrule
\multicolumn{1}{l|}{\multirow{6}{*}{\begin{tabular}[c]{@{}l@{}}Nominal lane \\follow \\ (968 scenarios)\end{tabular}}}                            & \multicolumn{1}{l|}{Average L2 error}                           & \multicolumn{1}{r|}{4.202058}    & 4.200189                     \\
\multicolumn{1}{l|}{}                                                                                                                 & \multicolumn{1}{l|}{Average L2 error 3s}                        & \multicolumn{1}{r|}{0.401251}    & 0.400440                     \\
\multicolumn{1}{l|}{}                                                                                                                 & \multicolumn{1}{l|}{Average L2 error 5s}                        & \multicolumn{1}{r|}{0.780286}    & 0.778495                     \\
\multicolumn{1}{l|}{}                                                                                                                 & \multicolumn{1}{l|}{Average L2 error 10s}                       & \multicolumn{1}{r|}{2.141165}    & 2.136004                     \\
\multicolumn{1}{l|}{}                                                                                                                 & \multicolumn{1}{l|}{Progress along expert route}                & \multicolumn{1}{r|}{0.939595}    & 0.939382                     \\
\multicolumn{1}{l|}{}                                                                                                                 & \multicolumn{1}{l|}{No at fault collisions}                     & \multicolumn{1}{r|}{0.909091}    & 0.909091                     \\ \midrule
\multicolumn{1}{l|}{\multirow{7}{*}{\begin{tabular}[c]{@{}l@{}}Decelerating from \\high speed \\ (1099 scenarios)\end{tabular}}}     & \multicolumn{1}{l|}{Average L2 error}                           & \multicolumn{1}{r|}{3.139028}    & 3.144618                     \\
\multicolumn{1}{l|}{}                                                                                                                 & \multicolumn{1}{l|}{Average L2 error 3s}                        & \multicolumn{1}{r|}{0.533925}    & 0.533905                     \\
\multicolumn{1}{l|}{}                                                                                                                 & \multicolumn{1}{l|}{Average L2 error 5s}                        & \multicolumn{1}{r|}{0.930384}    & 0.929786                     \\
\multicolumn{1}{l|}{}                                                                                                                 & \multicolumn{1}{l|}{Average L2 error 10s}                       & \multicolumn{1}{r|}{2.056039}    & 2.053457                     \\
\multicolumn{1}{l|}{}                                                                                                                 & \multicolumn{1}{l|}{Progress along expert route}                & \multicolumn{1}{r|}{0.990187}    & 0.990213                     \\
\multicolumn{1}{l|}{}                                                                                                                 & \multicolumn{1}{l|}{No at fault collisions}                     & \multicolumn{1}{r|}{0.945405}    & 0.942675                     \\
\multicolumn{1}{l|}{}                                                                                                                 & \multicolumn{1}{l|}{Deceleration time difference}               & \multicolumn{1}{r|}{0.443201}    & 0.439381                     \\ \midrule
\multicolumn{1}{l|}{\multirow{8}{*}{\begin{tabular}[c]{@{}l@{}}Start accelerating \\from stationary\\  (919 scenarios)\end{tabular}}} & \multicolumn{1}{l|}{Average L2 error}                           & \multicolumn{1}{r|}{8.919248}    & 9.019697                     \\
\multicolumn{1}{l|}{}                                                                                                                 & \multicolumn{1}{l|}{Average L2 error 3s}                        & \multicolumn{1}{r|}{0.426693}    & 0.425365                     \\
\multicolumn{1}{l|}{}                                                                                                                 & \multicolumn{1}{l|}{Average L2 error 5s}                        & \multicolumn{1}{r|}{1.244357}    & 1.249598                     \\
\multicolumn{1}{l|}{}                                                                                                                 & \multicolumn{1}{l|}{Average L2 error 10s}                       & \multicolumn{1}{r|}{4.611888}    & 4.673006                     \\
\multicolumn{1}{l|}{}                                                                                                                 & \multicolumn{1}{l|}{Progress along expert route}                & \multicolumn{1}{r|}{0.740881}    & 0.736437                     \\
\multicolumn{1}{l|}{}                                                                                                                 & \multicolumn{1}{l|}{No at fault collisions}                     & \multicolumn{1}{r|}{0.807399}    & 0.803047                     \\
\multicolumn{1}{l|}{}                                                                                                                 & \multicolumn{1}{l|}{Start from stationary max speed difference} & \multicolumn{1}{r|}{0.239367}    & 0.241552                     \\
\multicolumn{1}{l|}{}                                                                                                                 & \multicolumn{1}{l|}{Start from stationary time delay}           & \multicolumn{1}{r|}{3.477263}    & 3.494675                     \\ \bottomrule
\end{tabular}
\caption{Full nuPlan results comparing driving performance of our CW-Net planner to the original black-box ML planner.}
\label{tab:nuplan}
\end{table}






\begin{figure}[h!]
\centering
  \includegraphics[width=\textwidth]{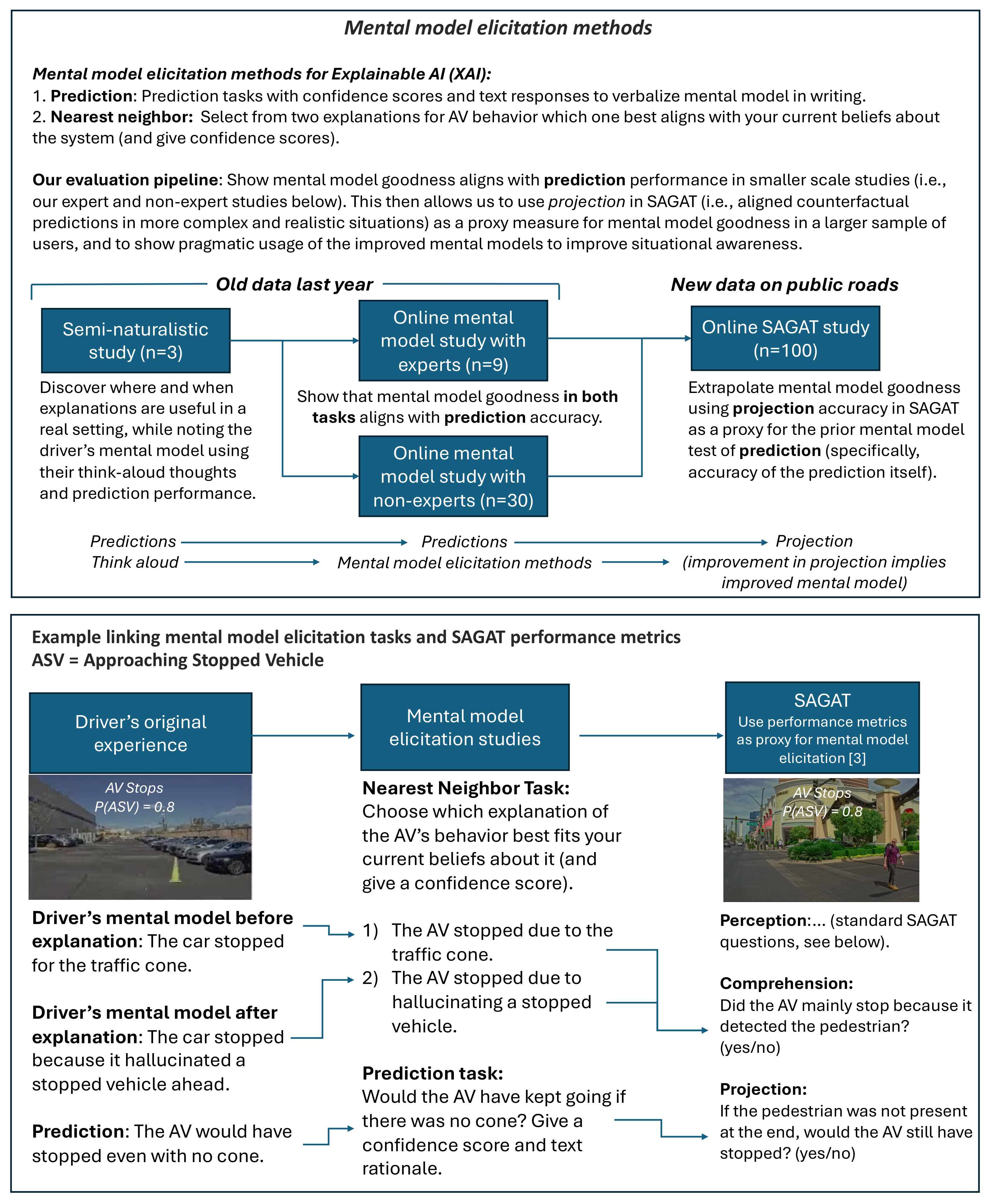}
  \caption{
  \textbf{Study methodology and pipeline example.}
  \textbf{a}, \textbf{Overall pipeline.} First we collected a series of surprising situations expert safety drivers encountered during their day-to-day job testing the AV where the explanations proved to be useful to understand why the AV made certain decisions and how it would behave in alternative scenarios. During these situations, we noted the driver's mental model through their think-aloud thoughts and behaviour. A mental model elicitation study was run with more expert drivers and non-experts online to more robustly verify these findings and whether mental model goodness did indeed correlate with counterfactual predictive performance. Then, in a large-scale SAGAT study, we used \textit{projection} as a surrogate metric for mental model quality, since it perfectly matches the prior prediction tasks.
  \textbf{b}, \textbf{Worked example.} with \texttt{ASV} concept.
  }
  \label{fig:studies_overview}
\end{figure}


\begin{figure}[h!]
\centering
  \includegraphics[width=\textwidth]{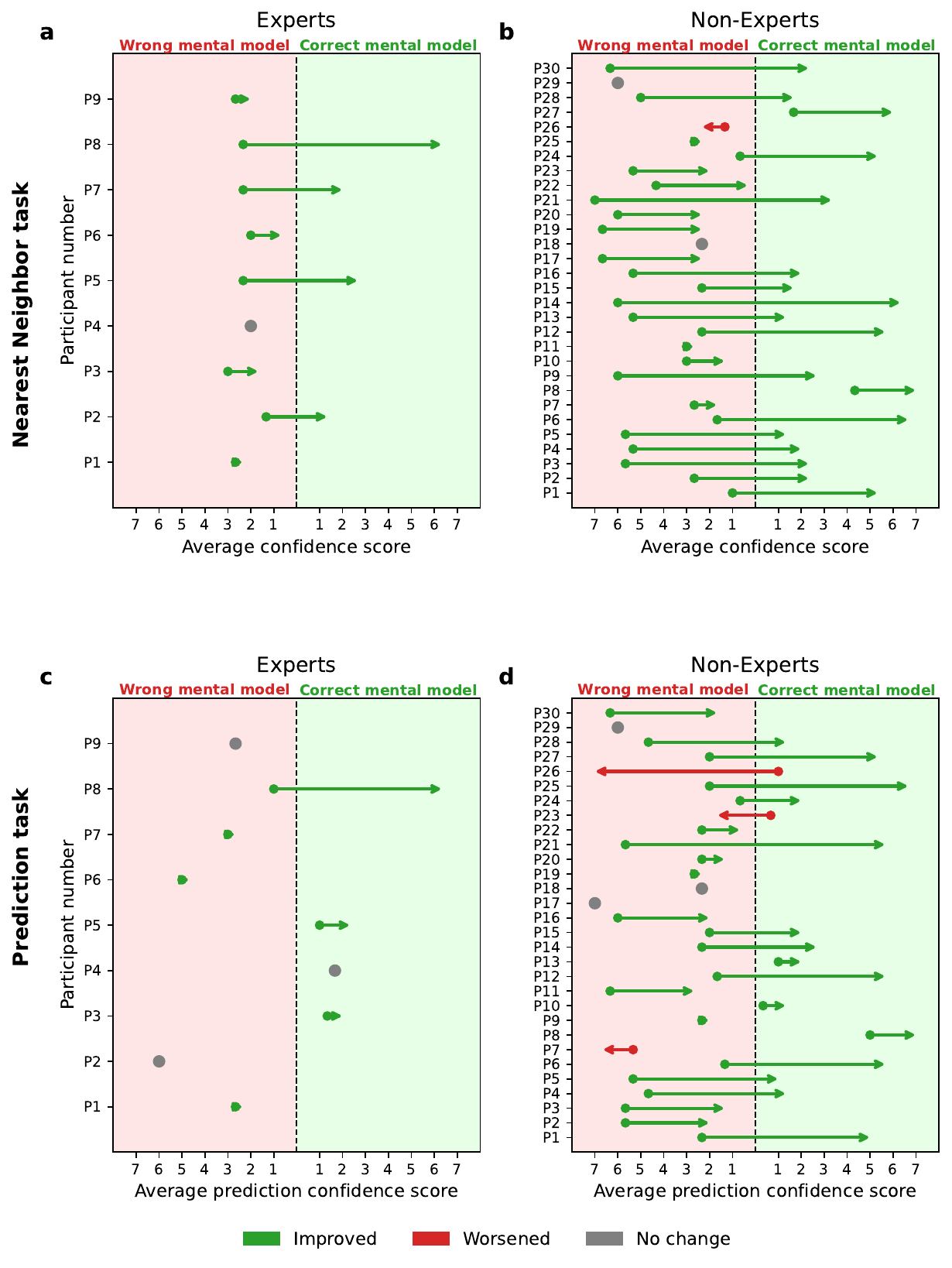}
  \caption{\textbf{Belief updates in response to explanations on the nearest-neighbor and prediction tasks.}
    Mental model improvement on the nearest-neighbor task (\textbf{a}, \textbf{b}) and performance improvement on the prediction task (\textbf{c}, \textbf{d}), for experts ($N=27$; \textbf{a}, \textbf{c}) and non-experts ($N=90$; \textbf{b}, \textbf{d}).
    Arrows originate at the participants' initial accuracy/confidence (before observing the explanation) and terminate at their final accuracy/confidence (after observing the explanation). For each participant, initial and final accuracy/confidence are averaged across all 3 scenarios. Green arrows indicate improvement in choice accuracy/confidence, red arrows indicate worsening, and gray points represent participants with no change in beliefs.}
  \label{fig:MM_combined}
\end{figure}


\begin{figure}[h!]
\centering
  \includegraphics[width=\textwidth]{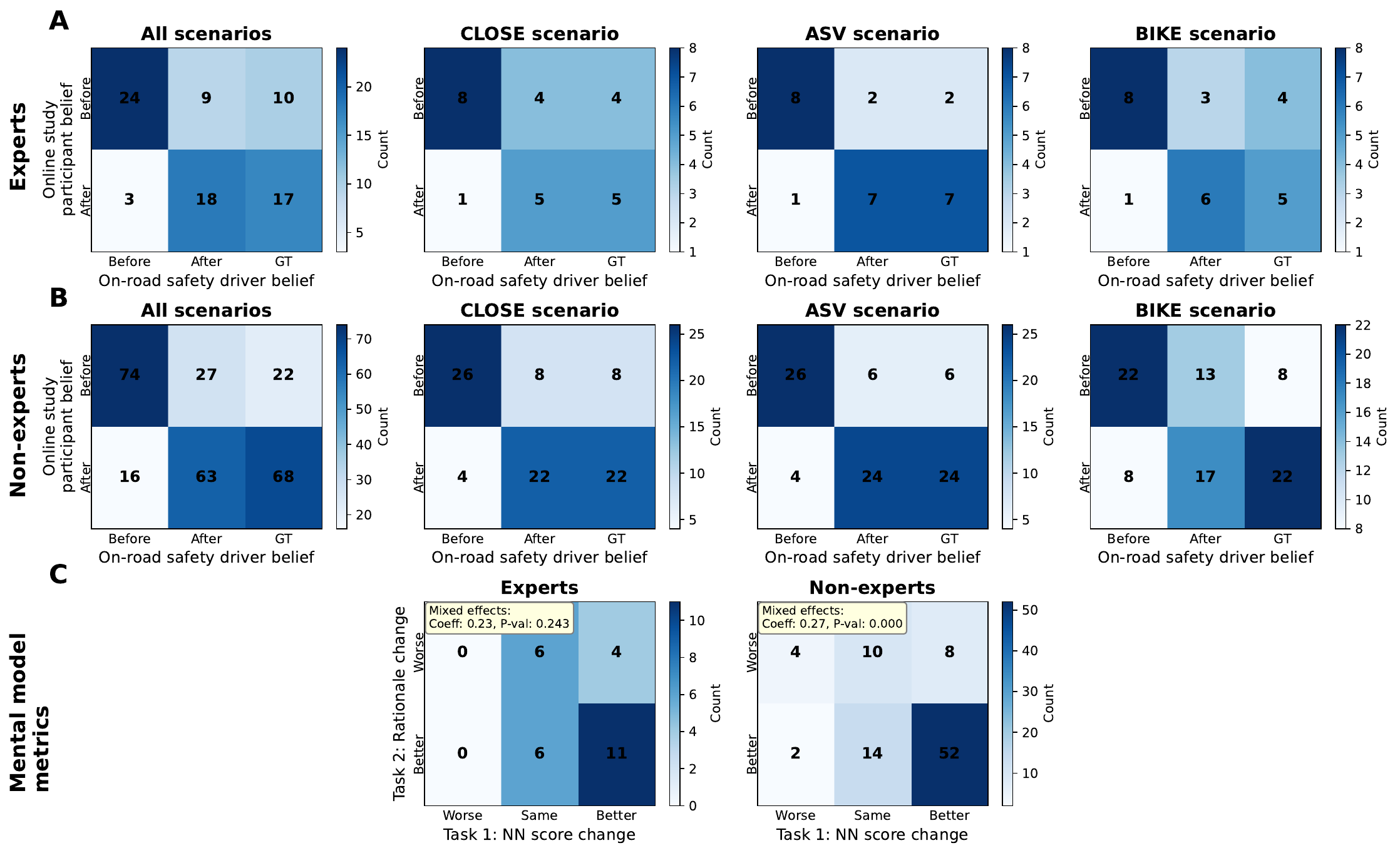}
  \caption{
  \textbf{A, B. Mental model improvement on free-form text rationale} for $N = 9$ experts (A) and $N = 30$ non-experts (B). Confusion matrices with participant beliefs (mental model and predictions) before/after observing the explanation, extracted from free-form responses in the mental model elicitation study study (y-axis), compared to beliefs from the safety drivers from the on-road tests before/after observing the explanation, as well as the ground truth (GT) explanation for AV behavior (x-axis). Numbers indicate raw response counts. \\
  \textbf{C. Correlation between mental model metrics across groups.} 
         Heatmaps displaying the frequency of alignment between changes in nearest-neighbor (NN) task responses and the free-form text rationale. To determine if the relationship between these two mental model proxies differs by expertise,  we performed ordinary least-squares (OLS) regression with clustered standard errors, including an interaction term. The interaction between group (experts vs. non-experts) and score change was not significant ($\beta = 0.04, p = 0.842$), indicating that the relationship between quantitative score improvements and qualitative rationale improvements is largely consistent across both groups.
  }
  \label{fig:llm_confusion_matrix__and__correlation_MM_heatmap}
\end{figure}

\begin{figure}[h!]
\centering
  \includegraphics[width=\textwidth]{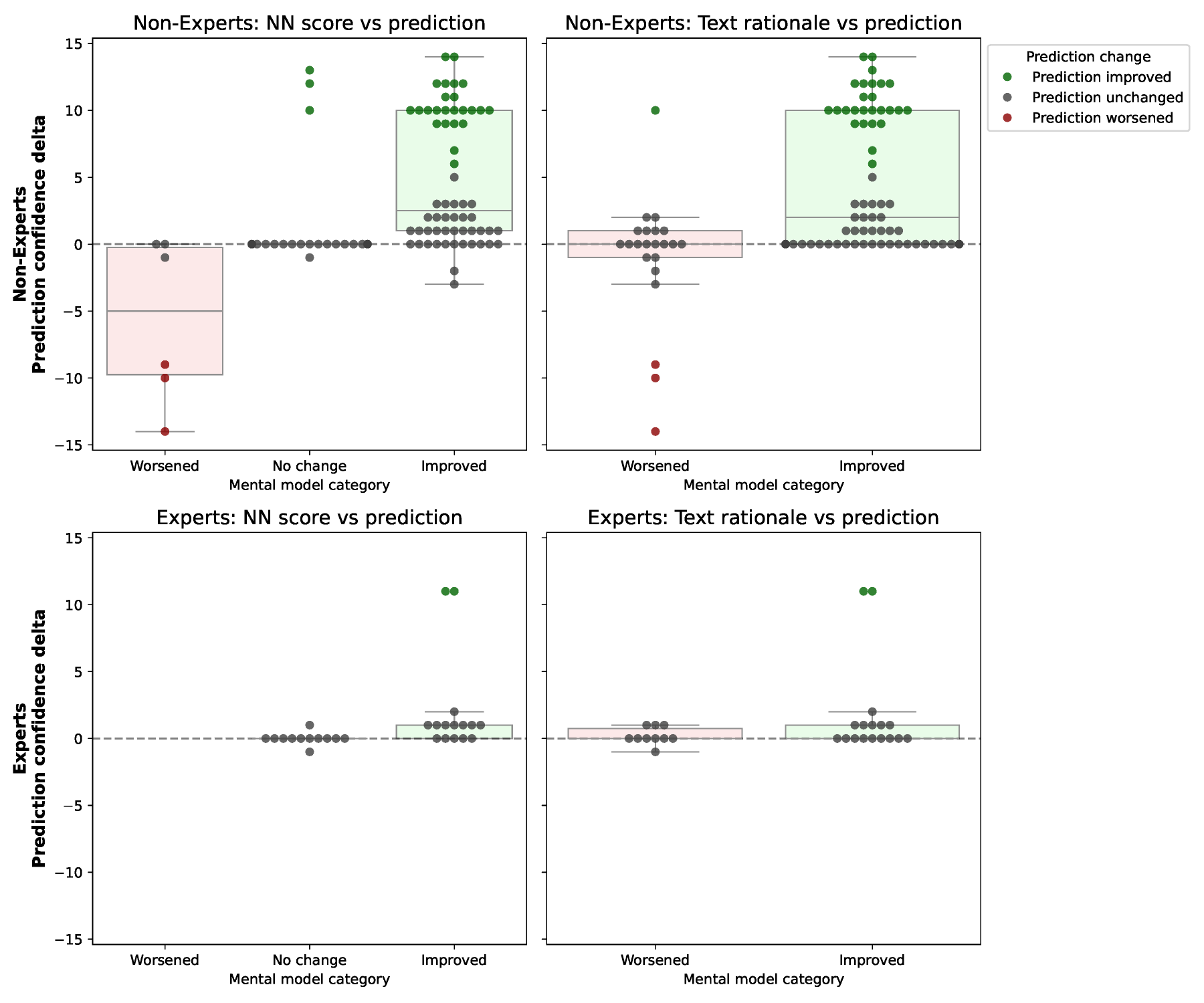}
    \caption{\textbf{Mental model improvement correlates with prediction improvement.} 
        Linear mixed effects (LME) models were used to analyze the relationship between changes in mental model proxies (nearest-neighbor task and free-form text rationale) and prediction accuracy/confidence. 
        Specifically, the y-axis indicates each participant's Likert confidence after explanation minus before, positive scores indicate they moved towards the correct prediction, and green dots indicate the prediction flipped correctly.
        For experts ($N=9$ participants x 3 scenarios), improvement in the nearest-neighbor (NN) task significantly correlates with prediction improvement ($\beta = 2.02, SE = 0.87, p = 0.02$). 
        The text rationale shows a similar positive effect ($\beta = 1.70, SE = 0.91, p = 0.06$), although it does not reach significance. 
        For non-experts ($N=30$ participants x 3 scenarios), both NN improvement ($\beta = 9.86, SE = 2.07, p < 0.001$) and text rationale improvement ($\beta = 5.03, SE = 1.23, p < 0.001$) are significantly correlated with prediction improvement. 
        Positive betas indicate that improving the mental model proxy is associated with a higher increase in prediction accuracy/confidence.
        }
    
  \label{fig:MM_cor_pred}
\end{figure}



\begin{figure}[h!]
\centering
  \includegraphics[width=\textwidth]{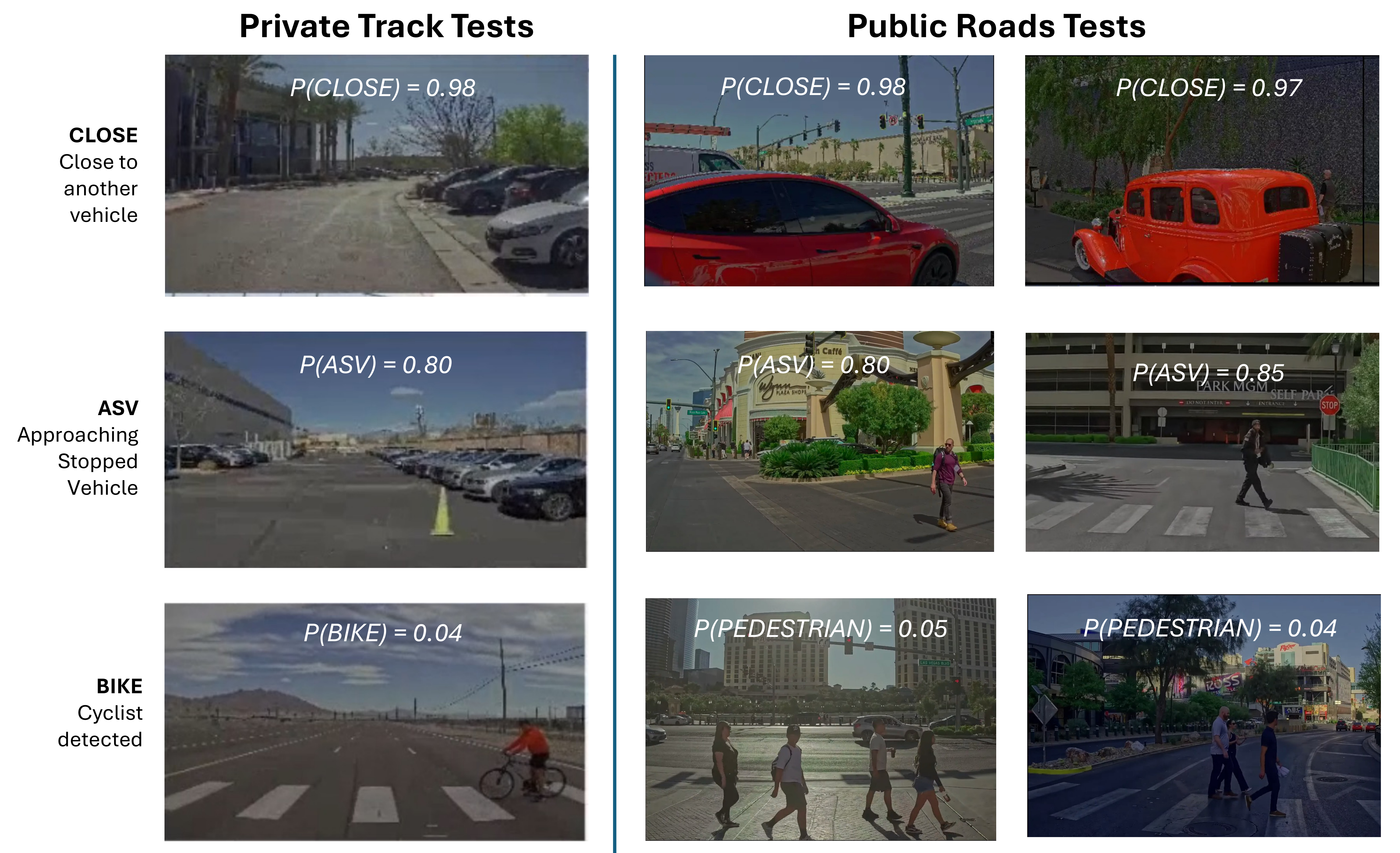}
  \caption{
  \textbf{Naturalistic scenario examples.} The original surprising scenarios discovered on the private track and the analogous scenarios discovered on public roads.
  }
  \label{fig:analogous_scenarios}
\end{figure}

\begin{figure}[h!]
\centering
  \includegraphics[width=\textwidth]{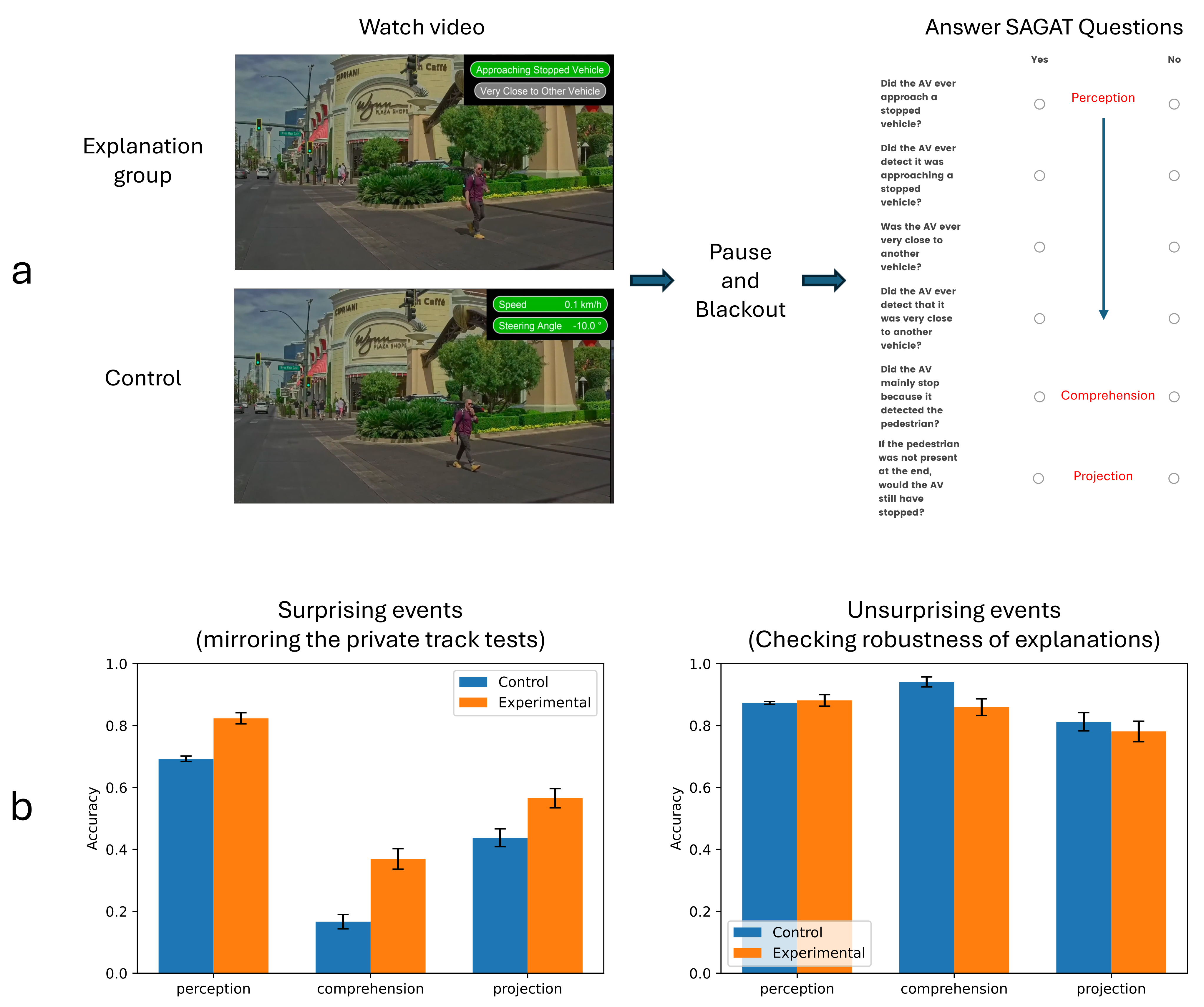}
  \caption{
  \textbf{a: SAGAT materials example.} 
   The two conditions (explanation vs. no explanation) were presented with the same video, followed by a ``blackout'', and then 6 questions related to situational awareness.
    \textbf{b: Online SAGAT study results}.
    Users in the experimental group reported significantly higher accuracy on perception, comprehension, and projection during surprising events. 
      During unsurprising events, there were no significant differences.
  Although a negative trend was noted for comprehension in the experimental group, it did not reach significance after Bonferroni correction (see Table~\ref{tab:survey_analysis}).
  Taken with the results of the mental model elicitation studies, which found counterfactual prediction accuracy correlated with mental model goodness, these SAGAT results strongly indicate the improvements in situational awareness were due to underlying improvements in user mental models of the AV.
  }
  
  \label{fig:SAGAT_example_and_results}
\end{figure}


\begin{figure}[h!]
\centering
  \includegraphics[width=\textwidth]{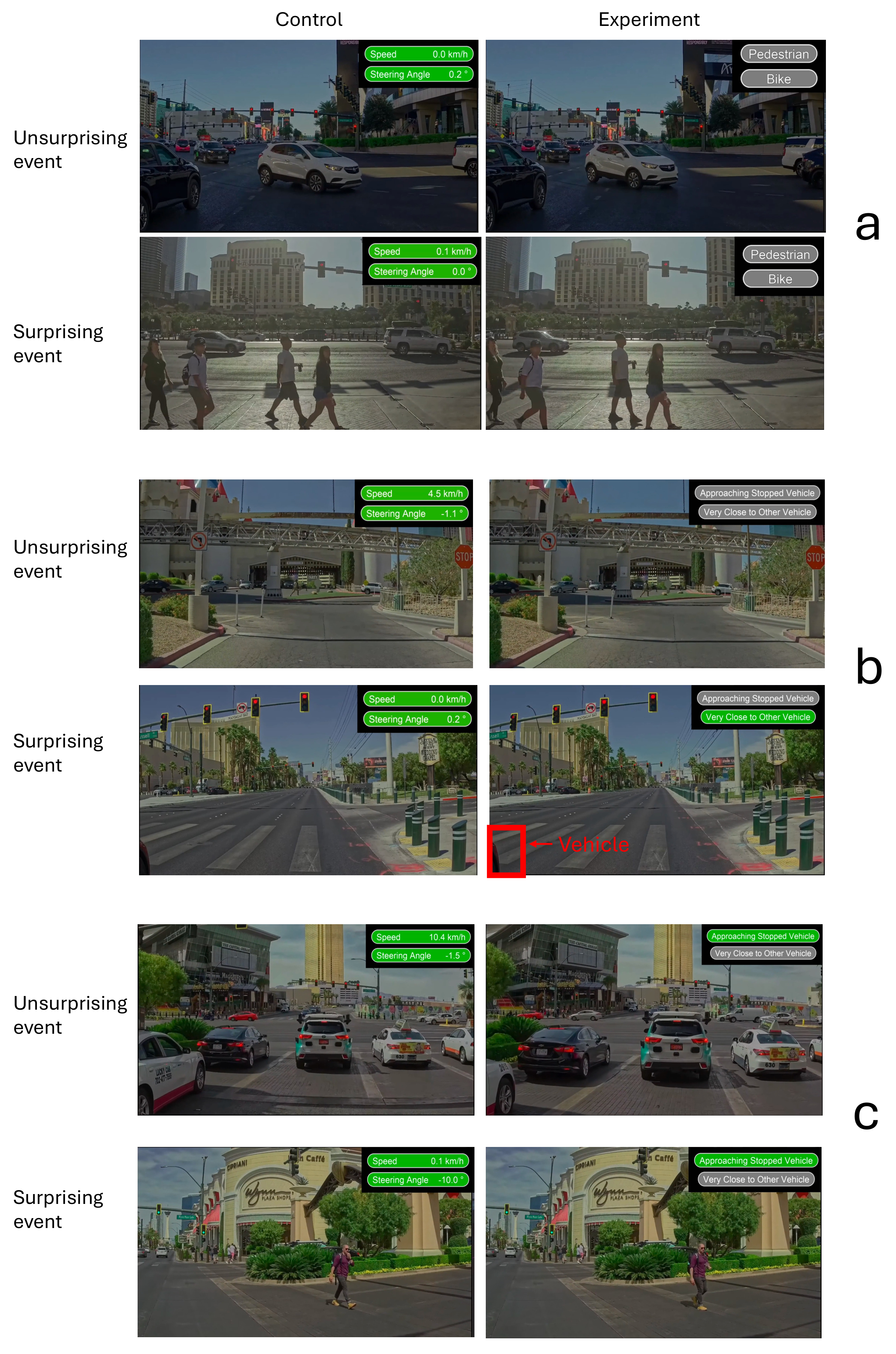}
  \caption{
  \textbf{User material examples for each concept.} Shown during a surprising and an unsurprising event, with the control condition on the left (speed and steering angle only) and the experiment condition on the right (concept labels overlaid). \textbf{a}, \texttt{PEDESTRIAN} concept (which replaced \texttt{BIKE} in our SAGAT study). \textbf{b}, \texttt{CLOSE} (very close to other vehicle) concept. \textbf{c}, \texttt{ASV} (approaching stopped vehicle) concept.
  }
  \label{fig:concepts_user}
\end{figure}

\clearpage

\section*{Supplementary Information}
\label{supplement}

\subsection*{Supplemental Methods}

\subsubsection*{Evaluating free-form human responses}

We used a large language model (LLM) as a cost-effective, reproducible, and unbiased way to semantically parse free-form text responses. We used the following prompts:

\begin{itemize}
    \item Instruction prompt: \textit{``Imagine you're a human participant in a psychological study. You are given a anchor mental model and prediction (denoted by ANCHOR) for the behavior of an autonomous vehicle (AV) and two candidate explanations and predictions for AV behavior (denoted by CANDIDATE 1 and CANDIDATE 2) given by passengers in the AV. Please indicate which of the two candidate explanations and predictions are most consistent with the anchor mental model and prediction. Respond simply with the number of the candidate (1 or 2).''}
    \item Input prompt:
    
        ``\textit{ANCHOR: $\langle$anchor$\rangle$}
        
        \textit{CANDIDATE 1:  $\langle$before$\rangle$}
        
        \textit{CANDIDATE 2:  $\langle$after$\rangle$}''
\end{itemize}

The input prompt was unique for every (participant, scenario, anchor) tuple. \textit{$\langle$before$\rangle$} and \textit{$\langle$after$\rangle$} were the participants' free-form responses before and after observing the explanation, respectively.  
There were three $\langle$anchor$\rangle$'s for each scenario, based on the private track tests and our post-hoc analyses (Figure~\ref{fig:results}) -- the safety driver's initial belief, the safety driver's final belief, and the ground truth reason for AV behavior:

\begin{itemize}
    \item CLOSE scenario:
    \begin{itemize}
        \item \textsc{Anchor 1} -- safety driver's initial belief: \textit{``The AV stopped because of the pickup/drop-off zone ahead. Therefore, if the AV was slightly to the left, it would not change its behavior; it would still stop.''},
        \item \textsc{Anchor 2} -- safety driver's final belief: \textit{``The AV stopped because of the parked cars nearby to the right. If it was further to the left, it would be further from the parked cars, and therefore it would keep moving.''},
        \item \textsc{Anchor 3} -- ground truth: \textit{``The AV stopped because of the parked cars nearby to the right. If it was further to the left, it would be further from the parked cars, and therefore it would keep moving.''},
    \end{itemize}
    \item ASV scenario:
    \begin{itemize}
        \item  \textsc{Anchor 1} -- safety driver's initial belief: \textit{``The AV stopped because of the traffic cone. Therefore, if the traffic cone was not there, it would have kept moving.''},
        \item  \textsc{Anchor 2} -- safety driver's final belief: \textit{``The AV stopped because it hallucinated a stopped vehicle, not because of the traffic cone. Therefore, if the traffic cone was not there, it may have still stopped.''},
        \item \textsc{Anchor 3} -- ground truth: \textit{``The AV stopped because it hallucinated a stopped vehicle, not because of the traffic cone. Therefore, if the traffic cone was not there, it would have still stopped.''},
    \end{itemize}
    \item BIKE scenario:
    \begin{itemize}
        \item \textsc{Anchor 1} -- safety driver's initial belief: \textit{``The AV can detect and stop for cyclists reliably, which is why it stopped for the cyclist. Therefore, we can trust that the AV will keep stopping reliably for cyclists in the future.''},
        \item \textsc{Anchor 2} -- safety driver's final belief: \textit{``The AV cannot detect and stop for cyclists reliably. Therefore, we cannot trust that the AV will keep stopping reliably for cyclists in the future.''},
        \item \textsc{Anchor 3} -- ground truth: \textit{``The AV cannot detect and respond to cyclists at all; instead, it stopped because of the automatic emergency braking system. Therefore, we cannot trust that the AV will keep stopping reliably for cyclists in the future.''},
    \end{itemize}
\end{itemize}

We used GPT-5, a state-of-the-art reasoning LLM. The prompts were tuned over 3 iterations on the expert responses (Figure~\ref{fig:llm_confusion_matrix__and__correlation_MM_heatmap}A) and then evaluated once on the non-expert responses (Figure~\ref{fig:llm_confusion_matrix__and__correlation_MM_heatmap}B). All code and data for this analysis will be available on \url{https://github.com/EoinKenny/CW_Net} upon publication.

To validate the judgments of the LLM, we performed an interrater reliability check by having two human raters act as judges for 6 of the non-expert participants' responses. The human judges were presented with the same prompts as the LLM with the goal of checking if the their judgements match those of the LLM. This resulted in a total of 54 human judgments (6 participants $\times$ 3 scenarios $\times$ 3 anchors) which were compared with each other and with the corresponding LLM judgments. 

Interrater reliability between the human judges was substantial (Cohen's $\kappa$ = 0.66), with agreement on 45 of the 54 (83\%) of judgments. The LLM had similar agreement with the two human judges: 85\% (46/54 judgments; Cohen's $\kappa$ = 0.70) with the first judge and 80\% (43/54 judgments; Cohen's $\kappa$ = 0.59) with the second judge. For the 45 queries on which the humans agreed with each other, the LLM achieved an even higher overall accuracy of $89\%$ (40/45 judgments; Cohen's $\kappa$ = 0.77). Together, these results indicate that the LLM judgments can serve as reliable substitutes for human judgments of the participants' free-form text responses.

\subsubsection*{Reliability of perception items}
\label{sec:reliability_perception}

We report the internal consistency of the SAGAT perception items and provide additional diagnostics to clarify how the composite perception score should be interpreted.

\paragraph{Measurement structure.}
We treat perception as a single latent construct measured by 48 binary indicators: 12 materials $\times$ 4 items per material. Each item probes a distinct concept. 
Across the full dataset, the resulting participant-by-item matrix has size $99 \times 48$. Because our main analyses use the aggregated composite perception score, reliability at this aggregated level is the relevant quantity for our manuscript.

\paragraph{Classical Cronbach's alpha.}
Computing Cronbach's alpha on the full $99 \times 48$ binary response matrix gives $\alpha = 0.754$, which is above the conventional 0.70 threshold for acceptable internal consistency (Table~\ref{tab:sagat_reliability}). For binary items, this calculation is equivalent to the KR-20 reliability coefficient. We therefore consider the full perception composite sufficiently reliable for the purposes of our analysis.

\paragraph{Interpretation for binary items.}
Alpha for binary correct/incorrect responses can be conservative when many items are near ceiling or floor. 
In such cases, limited item variance reduces the observed inter-item correlations, which can in turn depress alpha even when the items reflect a common underlying construct. This issue is relevant here because several of the world-state items were answered correctly by nearly all participants.

\paragraph{Item structure and ceiling effects.}
Our perception items divide naturally into two groups. World-state items (Q1 and Q3) ask what was present in the video, whereas AV-state items (Q2 and Q4) ask what the autonomous vehicle detected. The world-state items were closer to ceiling: pooled mean accuracy was $0.858$ for Q1 and $0.846$ for Q3 across $n=99$ participants, with 9 of 12 material-level Q1 items near ceiling, defined as accuracy $\geq 0.85$ (Table~\ref{tab:sagat_item_diagnostics}). By comparison, pooled mean accuracy was $0.786$ for Q2 and $0.785$ for Q4. The AV-state items also exhibited greater item-level response variance than the world-state items, making them less affected by ceiling restriction (Table~\ref{tab:sagat_item_diagnostics}).

\paragraph{Subscale reliability.}
For completeness, we also computed classical Cronbach's alpha separately for the AV-state and world-state subscales. The AV-state subscale achieves $\alpha = 0.792$, indicating good internal consistency. This is the component most directly targeted by the CW-Net explanations used in the main paper. The world-state subscale has a lower classical estimate, $\alpha = 0.408$, consistent with its substantially stronger ceiling restriction and lower item-level variance (Table~\ref{tab:sagat_reliability}).

\paragraph{Ordinal alpha.}
As a complementary reliability analysis for these binary items, we additionally report ordinal alpha. 
Ordinal alpha estimates reliability on an underlying latent response scale by using tetrachoric correlations rather than raw correlations between observed $0/1$ responses. 
This is appropriate when binary responses are viewed as discretized observations of an underlying continuous trait. Using this approach, ordinal alpha is $0.929$ for the full perception scale, $0.907$ for the AV-state subscale, and $0.862$ for the world-state subscale (Table~\ref{tab:sagat_reliability}). These values indicate that the binary-response format and associated ceiling effects likely make the classical alpha estimates conservative, particularly for the world-state items.

\paragraph{Excluded item for ordinal alpha.}
One item, an1--Q1, was excluded from the ordinal-alpha calculation because all 99 participants answered it correctly, resulting in zero variance. 
Since tetrachoric correlations are not estimable for an invariant item, the full-scale ordinal-alpha estimate is based on 47 of the 48 items, and the world-state ordinal-alpha estimate is based on 23 of the 24 items. The AV-state ordinal-alpha estimate uses all 24 AV-state items (Table~\ref{tab:sagat_reliability}).

\paragraph{Summary.}
Taken together, these analyses support the use of the composite perception score in the main text. 
Classical alpha on the full set of perception items is acceptable, the AV-state subscale is reliable on its own, and ordinal alpha shows that the binary-response format and ceiling effects likely attenuate the classical estimates.

\subsection*{Supplemental Tables}

\setcounter{table}{0}

\renewcommand{\thetable}{S\arabic{table}}

\begin{table}[!h]
\begin{tabular}{@{}lrcrr@{}}
\toprule
                             & \multicolumn{1}{c}{Accuracy} & Precision                & \multicolumn{1}{c}{Recall} & \multicolumn{1}{c}{F1 Score} \\ \midrule
Steering                     & 0.62                         & \multicolumn{3}{c}{(cross entropy)}                                                  \\
Speed                        & 0.83                         & \multicolumn{3}{c}{(cross entropy)}                                                  \\
Approaching stopped vechicle (ASV) & 0.55                         & \multicolumn{1}{r}{0.20}  & 0.89                       & 0.33                         \\
Intersection                 & 0.51                         & \multicolumn{1}{r}{0.42} & 0.61                       & 0.49                         \\
CLOSE to another vechicle    & 0.45                         & \multicolumn{1}{r}{0.10}  & 0.81                       & 0.17                         \\ \midrule
Ranker accuracy              & 0.94                         & \multicolumn{3}{c}{}                                                  \\ \bottomrule
\end{tabular}
\caption{CW-Net concept classification performance on Dataset 1 on holdout validation data.}
\label{table:concept500k}
\end{table}

\begin{table}[!h]
\begin{tabular}{@{}lrrrr@{}}
\toprule
                & \multicolumn{1}{c}{Accuracy} & \multicolumn{1}{c}{Precision} & \multicolumn{1}{c}{Recall} & \multicolumn{1}{c}{F1 Score} \\ \midrule
Slow            & 0.83                         & 0.73                          & 0.93                       & 0.82                         \\
Stopped         & 0.48                         & 0.16                          & 0.81                       & 0.27                         \\
Fast            & 0.68                         & 0.49                          & 0.86                       & 0.63                         \\
Stop sign       & 0.43                         & 0.03                          & 0.84                       & 0.05                         \\
Traffic light   & 0.39                         & 0.16                          & 0.62                       & 0.25                         \\
Intersection    & 0.42                         & 0.20                          & 0.65                       & 0.31                         \\
Pedestrian      & 0.44                         & 0.03                          & 0.84                       & 0.07                         \\
Following       & 0.43                         & 0.02                          & 0.84                       & 0.04                         \\
BIKE            & 0.21                         & 0.00                          & 0.42                       & 0.00                         \\
PUDO            & 0.58                         & 0.30                          & 0.86                       & 0.44                         \\ \midrule
Ranker accuracy & 0.95                         &                               &                            &                              \\ \bottomrule
\end{tabular}
\caption{CW-Net concept classification performance on Dataset 2 on holdout validation data.}
\label{table:concept3M}
\end{table}


\begin{table}[htbp]
    \centering
    \begin{tabular}{lcc}
        \toprule
        Scenario/concept & Before explanation & After explanation \\
        \midrule
        CLOSE & 0.00\% & 11.11\% \\
        ASV & 33.33\% & 44.44\% \\
        BIKE & 77.78\% & 77.78\% \\
        \midrule
        \textbf{Overall average} & 37.04\% & 44.44\% \\
        \bottomrule
    \end{tabular}
    \caption{Performance improvement on prediction task ($N = 9$ experts).
    }
    \label{tab:prediction_model_accuracy}
\end{table}

\begin{table}[htbp]
    \centering
    \begin{tabular}{lcc}
        \toprule
        Scenario/concept & Before explanation & After explanation \\
        \midrule
        CLOSE & 3.33\% & 33.33\% \\
        ASV & 36.67\% & 60.00\% \\
        BIKE & 40.00\% & 66.67\% \\
        \midrule
        \textbf{Overall average} & 26.67\% & 53.33\% \\
        \bottomrule
    \end{tabular}
    \caption{Performance improvement on prediction task ($N = 30$ non-experts). 
    }
    \label{tab:nonexperts_prediction_accuracy}
\end{table}



\begin{table}[]
\begin{tabular}{@{}lrrr@{}}
\toprule
\multicolumn{4}{c}{Surprising events}                                                                       \\ \midrule
      & \multicolumn{1}{c}{Concept activation} & \multicolumn{1}{c}{AV speed} & \multicolumn{1}{c}{Concept accuracy} \\ \midrule
\multicolumn{4}{c}{\textit{Original}}                                                                                \\ \midrule
ASV   & High                                   & Brakes                       & Bad                                  \\
CLOSE & High                                   & Frozen                       & Good                                 \\
BIKE  & Low                                    & Moving                       & Low                                  \\ \midrule
\multicolumn{4}{c}{\textit{Modified for unsurprising events}}                                                                 \\ \midrule
ASV   & High                                   & Brakes                       & \textbf{Good}                        \\
CLOSE & \textbf{Low}                           & Frozen                       & Good                                 \\
BIKE  & Low                                    & \textbf{Stopped}             & Low                                  \\ \bottomrule
\end{tabular}
\caption{Selecting unsurprising events for SAGAT study. To select unsurprising events we modifed what was unusual about each surprising event which fell into three categories.
\textbf{ASV}: This concept was surprising because of the misclassification of the concept, so we selected a scenario in which it accurately classified the concept.
\textbf{CLOSE}: This was surprising because the concept activation was higher than usual while the AV stopped, so we chose scenarios in which the AV stopped, but the concept activation was closer to its mean value.
\textbf{BIKE}: This was unusual because the AV stopped seemingly due to the cyclist, so we chose scenarios in which the AV again stopped for other reasons.
In all three setups, we only altered the one respective dimension, the other two remained static.}
\label{tab:non_surprising}
\end{table}

\begin{table}[ht]
\centering
\resizebox{\textwidth}{!}{%
\begin{tabular}{@{}llccccc@{}}
\toprule
\textbf{Valence} & \textbf{Dimension} & \textbf{Effect} & \textbf{Cohen's d} & \textbf{95\% CI} & \textbf{$p$-value} & \textbf{Bonferroni} \\ 
 & & \textbf{direction} & & & & \textbf{significant} \\ \midrule
Surprising events & Perception & Exp $>$ Control & 1.290 (large) & $[0.857, 1.723]$ & $< 0.001$ & Yes \\
 & Comprehension & Exp $>$ Control & 0.996 (large) & $[0.578, 1.413]$ & $< 0.001$ & Yes \\
 & Projection & Exp $>$ Control & 0.606 (medium) & $[0.203, 1.009]$ & 0.003 & Yes \\ \midrule
Unsurprising events & Perception & No effect & 0.085 (negligible) & $[-0.310, 0.479]$ & 0.667 & No \\
 & Comprehension & Control $>$ Exp & $-0.514$ (medium) & $[-0.915, -0.114]$ & 0.011 & No \\
 & Projection & No effect & $-0.142$ (negligible) & $[-0.536, 0.253]$ & 0.481 & No \\ \bottomrule
\end{tabular}%
}
\caption{SAGAT study results. During the `surprising' events, explanations had an all-around positive impact on people's situational awareness and by extension their mental models. Conversely, during `unsurprising' events, there were no significant differences between groups, after Bonferroni correction.}
\label{tab:survey_analysis}
\end{table}

\begin{table}[]
\begin{tabular}{@{}lccc@{}}
\toprule
\textbf{FEATURE} & \multicolumn{1}{r}{\textbf{Model 1}} & \multicolumn{1}{r}{\textbf{Model 2}} & \multicolumn{1}{r}{\textbf{Model 3}} \\ \midrule
STOPPED             & 0.0862                               & 0.0014                               & 0.0324                               \\
SLOW             & 0.1575                               & 0.0014                               & 0.0324                               \\
FAST             & 0.1903                               & —                                    & —                                    \\
STOP SIGN        & 0.2368                               & —                                    & —                                    \\
TRAFFIC LIGHT    & 0.0938                               & —                                    & —                                    \\
INTERSECTION     & 0.0856                               & 0.0247                               & 0.0535                               \\
PEDESTRIAN       & 0.0998                               & —                                    & —                                    \\
FOLLOWING        & 0.1056                               & —                                    & —                                    \\
BIKE             & 0.1222                               & —                                    & —                                    \\
PUDO             & 0.2347                               & —                                    & —                                    \\
LEFT             & —                                    & 0.0472                               & 0.0332                               \\
RIGHT            & —                                    & 0.0408                               & 0.0410                               \\
STRAIGHT         & —                                    & 0.0428                               & 0.0592                               \\
ASV              & —                                    & 0.0269                               & 0.1050                               \\
CLOSE            & —                                    & 0.0659                               & 0.0702                               \\ \bottomrule
\end{tabular}
\caption{Concept distribution shift: Wasserstein distances between the distribution of concept probabilities during the semi-naturalistic tests on the private track and the subsequent public-road test.
Overall, the differences varied from mostly insignificant to low/medium in \texttt{STOP SIGN} and \texttt{PUDO}.}
\label{tab:wasser}
\end{table}

\begin{table}[t]
\centering
\caption{\textbf{Internal-consistency estimates for the SAGAT perception items.}
Classical Cronbach's alpha was computed on the raw binary item matrix. Ordinal alpha was computed from the tetrachoric item-correlation matrix after excluding zero-variance items. The full-scale and world-state ordinal-alpha estimates exclude an1--Q1, which was answered correctly by all 99 participants.}
\label{tab:sagat_reliability}
\begin{tabular}{lccccc}
\toprule
Scale & $n$ & Items for classical $\alpha$ & Items for ordinal $\alpha$ & Classical $\alpha$ & Ordinal $\alpha$ \\
\midrule
Full perception scale (Q1--Q4) & 99 & 48 & 47 & 0.754 & 0.929 \\
AV-state subscale (Q2, Q4)     & 99 & 24 & 24 & 0.792 & 0.907 \\
World-state subscale (Q1, Q3)  & 99 & 24 & 23 & 0.408 & 0.862 \\
\bottomrule
\end{tabular}
\end{table}

\begin{table}[t]
\centering
\caption{\textbf{Response-distribution diagnostics for the four SAGAT perception item types.}
Pooled mean accuracy is computed across all 99 participants and 12 materials for each question type. Mean item variance is the average variance across the 12 material-level binary indicators for that question. ``Near ceiling'' denotes material-level items with accuracy $\geq 0.85$.}
\label{tab:sagat_item_diagnostics}
\begin{tabular}{llccc}
\toprule
Question & Item type & Pooled mean accuracy & Mean item variance & Near-ceiling items \\
\midrule
Q1 & World-state & 0.858 & 0.094 & 9/12 \\
Q2 & AV-state    & 0.786 & 0.130 & 7/12 \\
Q3 & World-state & 0.846 & 0.105 & 7/12 \\
Q4 & AV-state    & 0.785 & 0.124 & 7/12 \\
\bottomrule
\end{tabular}
\end{table}

\newpage

\end{document}